\pdfoutput=1

\documentclass[11pt]{article}

\PassOptionsToPackage{table}{xcolor}
\usepackage[preprint]{acl}

\usepackage{times}
\usepackage{latexsym}
\usepackage[T1]{fontenc}
\usepackage[utf8]{inputenc}
\usepackage{microtype}
\usepackage{inconsolata}
\usepackage{graphicx}

\usepackage{url}
\usepackage{array}
\usepackage{booktabs}
\usepackage{amsfonts}
\usepackage{amsmath}
\usepackage{amssymb}
\usepackage{amsthm}
\usepackage{nicefrac}
\usepackage{multirow}
\usepackage{enumitem}
\usepackage{xspace}
\usepackage{float}
\usepackage{algorithm}
\usepackage{algpseudocode}
\usepackage{dsfont}

\usepackage{tcolorbox}
\tcbuselibrary{breakable, skins}
\definecolor{deepgreen}{RGB}{0,100,0}
\definecolor{deeppurple}{RGB}{120,0,120}
\definecolor{deepblue}{RGB}{0,40,140}
\definecolor{deeporange}{RGB}{200,80,0}

\newtheorem{proposition}{Proposition}

\newcommand{\ind}{\mathds{1}}
\newcommand{\method}{GCA\xspace}
\newcommand{\toolhit}{\textsc{ToolHit}}
\newcommand{\hitacc}{\textsc{HitAcc}}

\newcommand{\supcorr}{\textsc{SupCor}}
\newcommand{\unsupcorr}{\textsc{UnsupCor}}
\newcommand{\valids}{\textsc{vSearch}}
\newcommand{\redunds}{\textsc{rSearch}}
\newcommand{\answ}{\textsc{Ans}}
\newcommand{\Retr}{\mathrm{Retr}}

\title{Grounded Credit Assignment in Reinforcement Learning
for Multi-Turn Search Agents}

\author{%
  \textbf{Qirui Liu\textsuperscript{1*}},
  \textbf{Hao Chen\textsuperscript{2*\dag}},
  \textbf{Yi Wen\textsuperscript{3}},
  \textbf{Weijie Shi\textsuperscript{4}},
  \textbf{Yao Zhao\textsuperscript{5}},
  \\
  \textbf{Xianhe Chen\textsuperscript{6}},
  \textbf{Hao Zhao\textsuperscript{2}},
  \textbf{Xing Tang\textsuperscript{7}},
  \textbf{Weihong Luo\textsuperscript{2}},
  \textbf{Yuyu Luo\textsuperscript{8\dag}}
  \\
  \fontsize{10pt}{12pt}\selectfont
  \begin{tabular}[t]{@{}c@{}}
    \textsuperscript{1}South China University of Technology,
    \textsuperscript{2}FiT, Tencent,
    \textsuperscript{3}City University of Hong Kong,
    \\
    \textsuperscript{4}The Hong Kong University of Science and Technology,
    \textsuperscript{5}Nanjing University,
    \textsuperscript{6}Imperial College London,
    \\
    \textsuperscript{7}Shenzhen Technology University,
    \textsuperscript{8}The Hong Kong University of Science and Technology (Guangzhou)
    \\
    \textsuperscript{*}Equal contribution. \quad \textsuperscript{\dag}Corresponding authors.
  \end{tabular}
}

\begin{document}

\maketitle

\begin{abstract}
Outcome-reward reinforcement learning such as GRPO gives every token of a search-agent rollout the same advantage, although the rollout exercises two capabilities: finding evidence and using it.
We show that accuracy improves without grounding: Search-R1 raises accuracy on all seven QA benchmarks, yet on Bamboogle the evidence-supported share of correct answers falls from 84.0\% to 42.9\%.
We attribute this to GRPO's credit assignment: rollouts that retrieve answer-bearing evidence are favoured in aggregate only in proportion to their accuracy gain over the rest, a \emph{grounding gap} that answering from memory or misusing evidence shrinks or reverses.
To address this, we propose Grounded Credit Assignment (\method\footnote{Code: \url{https://anonymous.4open.science/r/Grounded-Credit-Assignment-6320}}), which uses verifiable sub-outcomes (evidence retrieval and answer correctness) to correct search and answer credit separately: a wrong answer is blamed more than the search that found the evidence, and a correct answer that bypasses retrievable evidence is credited less.
Never negative on evidence-finding search, the correction does not shrink with the grounding gap; capped at 10\% of each token's advantage, it never overrides the outcome signal.
Without extra models, rollouts, or training stages, \method improves average EM over Search-R1 by 5.7 (Qwen2.5-3B) and 6.3 points (Qwen3-30B-A3B) and raises this share from 76.9\% to 93.4\%.
\end{abstract}

\section{Introduction}
\label{sec:intro}

Reinforcement learning with verifiable outcome rewards~\citep{deepseekai2025r1,yu2025dapo} has become the dominant recipe for training multi-turn search agents that interleave reasoning, tool calls, and answer generation~\citep{jin2025searchr1,song2025r1searcher,zheng2025deepresearcher}.
In the Search-R1 recipe~\citep{jin2025searchr1}, a group of rollouts is sampled for each question, each rollout is scored by whether its final answer matches the gold answer, and the group-relative advantage is broadcast to every assistant token.
A search-agent rollout, however, exercises two different capabilities: \emph{acquiring} evidence through search and \emph{using} it to answer.
A single scalar cannot tell which of the two produced the outcome.

\begin{figure*}[t]
\centering
\includegraphics[width=\textwidth]{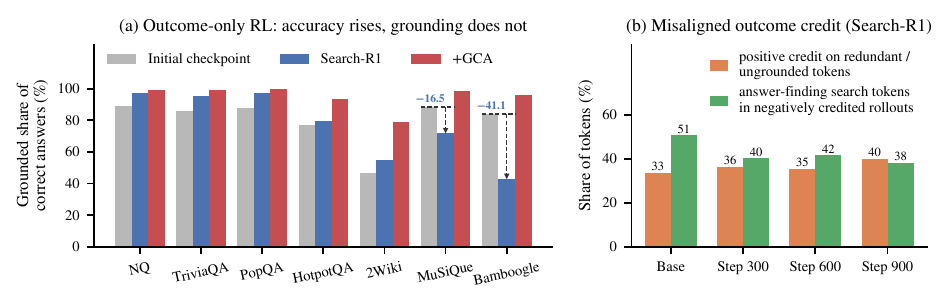}
\caption{\textbf{Outcome-only RL improves accuracy without improving grounding.}
(a)~Share of correct answers that are supported by retrieved evidence (\supcorr/Acc) on Qwen2.5-3B. Search-R1 raises EM on every benchmark (Table~\ref{tab:main-em}) but lowers this share below the initial checkpoint (before task RL) on MuSiQue and Bamboogle; \method restores it.
(b)~Token-level view of the Search-R1 run (Appendix~\ref{app:eda-observations}): share of positively credited tokens (type $A$ or $D$ rollouts) that are post-hit redundant searches or belong to ungrounded type-$D$ rollouts, and share of answer-finding search tokens that lie in hit-but-wrong (type $B$) rollouts. Shares are computed per rollout type and ignore homogeneous groups, whose advantage is zero.}
\label{fig:motivation}
\end{figure*}

This is not a benign simplification.
Figure~\ref{fig:motivation}a shows that Search-R1 raises exact match on all seven benchmarks we study, yet on MuSiQue and Bamboogle the share of its correct answers that are supported by retrieved evidence falls \emph{below} that of the initial checkpoint (88.3\%$\to$71.8\% and 84.0\%$\to$42.9\%): the agent increasingly answers from parametric memory.
We attribute this drift to the single scalar.
Under GRPO, the net advantage of the rollouts in a group that retrieved answer-bearing evidence equals a positive factor times the group's \emph{grounding gap}, the accuracy difference between rollouts with and without such evidence (Proposition~\ref{prop:gap}).
In aggregate, retrieval is thus credited only insofar as it coincides with correct \mbox{answers}.

The token-level view shows how this credit erodes (Figure~\ref{fig:motivation}b): 38--51\% of the search tokens up to the first answer-bearing hit lie in rollouts that later answered incorrectly, so their negative credit offsets what such searches earn in successful rollouts, while 33--40\% of positively credited tokens are redundant searches or belong to ungrounded rollouts.
When the model can guess from memory, or fails to use the evidence it found, the gap shrinks or reverses, and the aggregate advantage no longer favours rollouts that found evidence, consistent with Figure~\ref{fig:motivation}a.
We call this \emph{credit conflation}.

In search, however, each capability has its own verifiable sub-outcome: whether a tool response contains the gold answer, and whether the final answer matches it.
We propose \emph{Grounded Credit Assignment} (\method), which adds to each functional segment a residual whose polarity follows the sub-outcome the segment controls.
\method thus shifts part of the blame for a wrong answer from the search that found the evidence to the answer, and reduces the credit of redundant searches and of correct answers that bypass retrievable evidence.
Bounded so that no token changes the sign of its advantage (Proposition~\ref{prop:sign}), the residual reweights outcome credit by at most 10\% rather than overriding it.
A correction this small can still matter: it is never negative on valid search, so, unlike the outcome credit, it does not cancel across rollouts.
A margin--entropy salience score concentrates the residual on decision-critical tokens, and \method needs no auxiliary model, extra rollouts, or staged training.
Our contributions are:
\begin{itemize}[leftmargin=1.2em,topsep=2pt,itemsep=1pt,parsep=0pt]
\item We identify and formalise credit conflation in outcome-reward search RL, through a grounding-gap identity and token-level evidence of misaligned credit.
\item We propose \method, which corrects each segment's credit by the verifiable sub-outcome it controls (via a first-hit proxy) with a provably sign-preserving residual whose magnitudes can be learned without loss.
\item On seven QA benchmarks, \method improves average EM over Search-R1 by 5.7 (Qwen2.5-3B, GRPO) and 6.3 points (Qwen3-30B-A3B, GSPO) and raises the evidence-supported share of correct answers from 76.9\% to 93.4\%; ablations, learned routing, and an audit of the grounding signal show that the gains do not hinge on hand-tuned rules.
\end{itemize}

\section{Related Work}
\label{sec:related}

\paragraph{RL for search agents.}
Search-R1~\citep{jin2025searchr1} trains LLMs to interleave reasoning and retrieval with an exact-match reward and retrieved-token masking.
Follow-up work scales the environment to the web~\citep{song2025r1searcher,zheng2025deepresearcher}, simulates the search engine~\citep{sun2025zerosearch}, trains only the searcher~\citep{jiang2025s3}, adds evidence-quality rewards~\citep{zhao2025rsearch,he2026searchr2}, or branches rollouts at high-entropy tool calls~\citep{dong2025arpo}.
These methods change the environment, the reward, or the rollout procedure, but still assign one advantage to all tokens of a trajectory or turn.
\method keeps all three fixed and changes only how the advantage is distributed over tokens.

\paragraph{Finer-grained credit assignment.}
DeSA~\citep{wang2025desa} decouples search and answering with a two-stage curriculum, which breaks end-to-end training and requires choosing a stage-switch checkpoint.
IGPO~\citep{wang2026igpo} rewards each turn by its information gain, MT-PPO~\citep{wei2025turnlevel} adds verifiable turn-level rewards and a critic, and GiGPO~\citep{feng2025group} and HGPO~\citep{he2026hierarchy} compute step-level advantages by grouping actions that share states or histories.
These methods group credit by time step or environment state; \method groups it by the \emph{functional role} of a segment, an orthogonal axis.

\paragraph{Token-level weighting.}
\citet{wang2025highentropy} restrict policy gradients to high-entropy ``forking'' tokens, and RTT~\citep{xu2026rtt} learns a token-level relevance discriminator for instruction following.
\method uses margin and entropy only to distribute a segment residual whose sign is already fixed by the diagnosis, and requires no extra model.
Appendix~\ref{app:positioning} compares the design choices of these methods.

\section{Credit Conflation in Outcome-Reward Search RL}
\label{sec:conflation}

\subsection{Setup}
\label{sec:prelim}

For a question $x$, the policy $\pi_{\theta_{\mathrm{old}}}$ samples a group $\mathcal{G}(x)=\{r_1,\ldots,r_n\}$ of multi-turn rollouts.
In each rollout the assistant calls a search tool through structured function calling, and the $k$-th call returns a tool response $o_{i,k}$ with the top-$K$ passages.
Following Search-R1, only assistant tokens are optimised.
Each rollout receives an outcome reward $R_i$ (exact match plus a format term) and the group-relative advantage $\hat A_i=(R_i-\mu_{\mathcal G})/\sigma_{\mathcal G}$ of GRPO~\citep{shao2024deepseekmath} or GSPO~\citep{zheng2025gspo}, which is broadcast to all of its tokens.

\paragraph{Verifiable sub-outcomes.}
Let $\mathcal{A}(x)$ be the gold alias set and $m(a,o)=\ind[\mathrm{lower}(\mathrm{strip}(a))\subseteq\mathrm{lower}(o)]$ a case-insensitive containment test.
We define
\begin{equation}
\label{eq:subout}
\begin{aligned}
H_i &= \max\nolimits_{k}\max\nolimits_{a\in\mathcal{A}(x)} m(a,o_{i,k}),\\
k^\star_i &= \min\{k:\max\nolimits_{a\in\mathcal{A}(x)} m(a,o_{i,k})=1\},\\
C_i &= \max\nolimits_{a\in\mathcal{A}(x)}\ind[N(\hat y_i)=N(a)],
\end{aligned}
\end{equation}
where $\hat y_i$ is the final answer and $N(\cdot)$ is the standard EM normalisation.
$H_i$ (\toolhit) records whether the agent ever \emph{saw} answer-bearing evidence, $k^\star_i$ is the first search turn that did, and $C_i$ records correctness.
The pair $(H_i,C_i)$ yields four rollout types: $A$ ($H{=}1,C{=}1$, grounded success), $B$ ($H{=}1,C{=}0$, evidence found but misused, $B_1$, or the turn budget exhausted, $B_2$), $C$ ($H{=}0,C{=}0$), and $D$ ($H{=}0,C{=}1$, unsupported correctness).
For $H{=}0$ rollouts we also test whether the corpus contains the answer at all, $\Retr(x)$, by issuing templated oracle queries built from the gold aliases; $\Retr(x)$ is set to $1$ whenever any rollout of the group has $H_i{=}1$ (Appendix~\ref{app:retrievability}).

\subsection{The grounding-gap identity}
\label{sec:gap}

\begin{proposition}[Grounding-gap identity]
\label{prop:gap}
Consider a group of $n$ rollouts with binary rewards $R_i=C_i$ and reward standard deviation $\sigma>0$. Let $p_H$ be the fraction of rollouts with $H_i{=}1$, where $0<p_H<1$, and $\Delta=\Pr[C{=}1\mid H{=}1]-\Pr[C{=}1\mid H{=}0]$ the within-group grounding gap. Then
\begin{equation}
\sum_{i:\,H_i=1}\hat A_i = -\!\!\sum_{i:\,H_i=0}\hat A_i = \frac{np_H(1-p_H)}{\sigma}\Delta .
\end{equation}
\end{proposition}

The proof takes three lines (Appendix~\ref{app:proofs}); its covariance form also covers a format term and $p_H\in\{0,1\}$, where $\Delta$ is undefined and the sum is zero.
The identity concerns aggregate scalar advantages, not the full search-token policy gradient, which also depends on token-level score-function gradients.
We draw three observations.

\emph{(i) Vanishing aggregate credit.}
In aggregate, hit rollouts are favoured over miss rollouts only through $\Delta$.
Parametric shortcuts raise $\Pr[C\mid H{=}0]$ (type $D$) and evidence misuse lowers $\Pr[C\mid H{=}1]$ (type $B$); both shrink $\Delta$ and can make it negative, in which case the rollouts that found the evidence are penalised in aggregate.
This is consistent with Search-R1 lowering \toolhit{} on MuSiQue (25.8$\to$21.9) and Bamboogle (35.2$\to$19.2) relative to the initial checkpoint (Appendix~\ref{app:detail}).

\emph{(ii) Split credit inside hit rollouts.}
Even when $\Delta>0$, the credit of hit rollouts is a sum of positive type-$A$ and negative type-$B$ advantages, so the search turns up to the first hit in a type-$B$ rollout are pushed down together with its wrong answer (38--51\% of answer-finding search tokens; Figure~\ref{fig:motivation}b).

\emph{(iii) Conflict cells.}
We call a segment a conflict cell when its outcome sign disagrees with the sign of the sub-outcome it controls: valid search in type $B$ (outcome $-$, acquisition $+$), redundant search after the first hit in type $A$ (outcome $+$, no acquisition), and both segments of type $D$ (outcome $+$, failed acquisition and an ungrounded answer).
Table~\ref{tab:routing} marks these cells.

\section{Grounded Credit Assignment}
\label{sec:method}

\begin{figure*}[t]
\centering
\includegraphics[width=\textwidth]{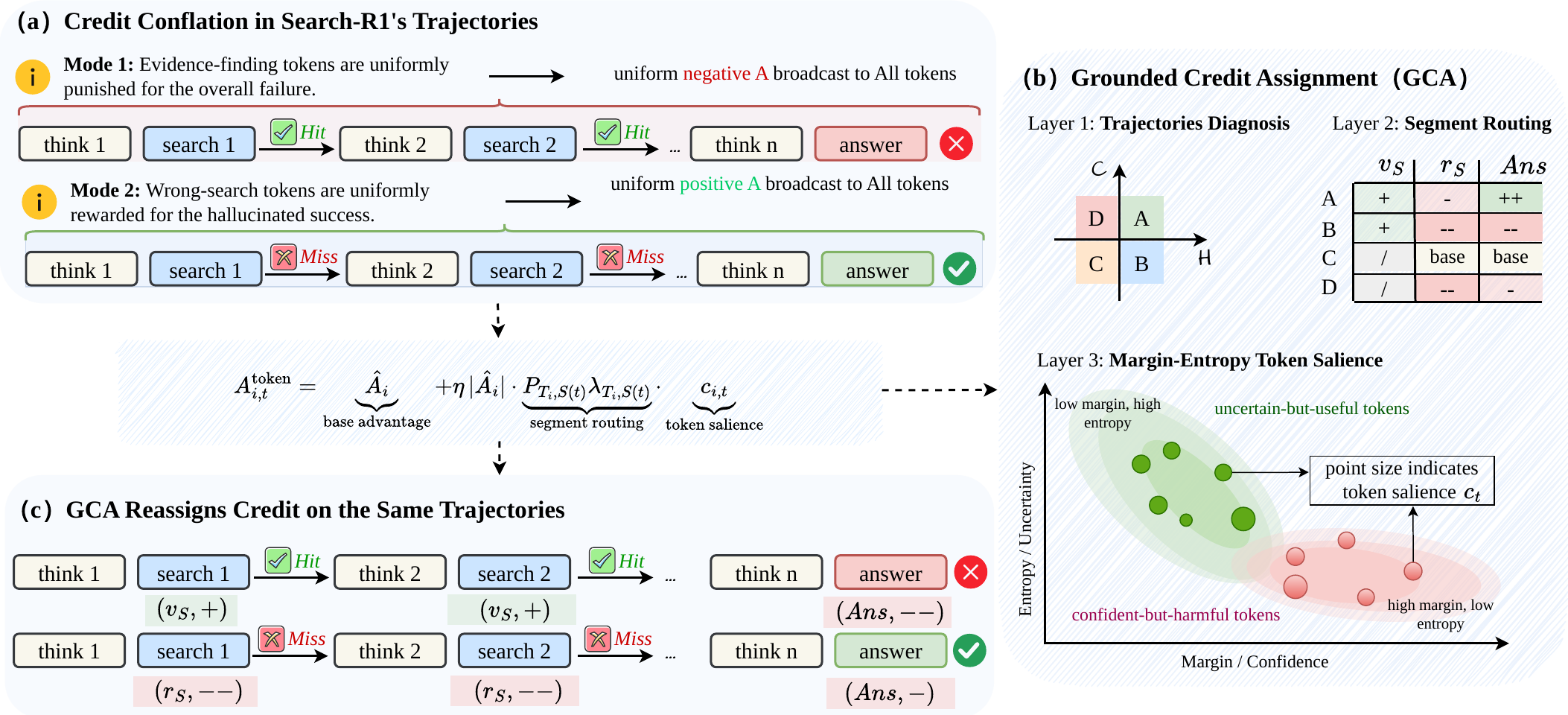}
\caption{\textbf{Overview of \method.} (a)~Uniform advantage broadcasting conflates search and answer credit within a rollout. (b)~\method diagnoses each rollout by retrieval hit $H$ and answer correctness $C$, routes signed residuals to functional segments (\valids, \redunds, \answ), and concentrates them on decision-critical tokens via margin--entropy salience. (c)~Resulting residuals; signs show residual polarity, not the final advantage's sign.}
\label{fig:gca-method}
\end{figure*}

Guided by this analysis, \method corrects the token-level advantage, leaving the reward, the group statistics, and the optimiser unchanged (Figure~\ref{fig:gca-method}).

\subsection{Routing motivated by verifiable trajectory sub-outcomes}
\label{sec:routing}

Each assistant token belongs to one functional segment $S(t)$: \valids{} contains the valid search turns up to the first hit ($k\le k^\star_i$); \redunds{} contains later searches, all searches of $H{=}0$ rollouts, and malformed calls; and \answ{} contains the final-answer turn.
This first-hit segmentation is an operational proxy for acquisition-related and potentially redundant search; it does not establish the causal contribution of every turn.
The \emph{sub-outcome} of a segment is the part of the result it controls.
Under this proxy, \valids{} acquired the evidence (acquisition $+$) and \redunds{} did not (acquisition $-$); the answer used available evidence correctly ($+$, type $A$), misused it ($-$, type $B$), or bypassed retrievable evidence ($-$, type $D$ with $\Retr{=}1$).
The residual polarity is set to this sign,
\begin{equation}
P_{T,S}=\mathrm{sign}\bigl(\text{sub-outcome of } S \text{ in type } T\bigr),
\end{equation}
a proxy-label rule that gives the signs in Table~\ref{tab:routing}, rather than a unique or optimal assignment implied by Proposition~\ref{prop:gap}.
Two rows need no residual.
In type $C$, the outcome and the sub-outcomes are all negative or undefined, so there is no conflict to correct; a residual there would only add negative pressure on the largest token mass, the kind of pressure that destabilises training (Appendix~\ref{app:calibration}).
In type $D$ with $\Retr{=}0$ the corpus lacks the answer, so a parametric answer is legitimate and the rollout is masked from the policy loss; it still enters the group statistics, and groups consisting only of such rollouts are below 0.1\%.

The magnitudes $\lambda_{T,S}\in[0,1]$ follow a bottleneck rule: within each type, the segment that decided the outcome receives the largest magnitude, useful search inside a failed rollout keeps a moderate positive residual, and the outcome credit of a correct but ungrounded answer is only softly downweighted (Appendix~\ref{app:per-segment}).
Section~\ref{sec:learned-routing} shows that learning these magnitudes under the same budget does not improve results.

\begin{table}[t]
\caption{\textbf{Routing coefficients.} Entries are signed residual coefficients $P_{T,S}\lambda_{T,S}$; signs follow each segment's proxy sub-outcome (\S\ref{sec:routing}). Shaded: conflict cells, where outcome advantage and sub-outcome disagree in sign. \textsc{Base}: original advantage only; \textsc{Mask}: excluded from the policy loss; --: segment absent.}
\label{tab:routing}
\centering
\small
\setlength{\tabcolsep}{3pt}
\begin{tabular}{@{}llccc@{}}
\toprule
Type & Condition & \valids & \redunds & \answ \\
\midrule
$A$   & $H{=}1,C{=}1$          & $+0.5$ & \cellcolor{gray!22}$-0.1$ & $+1.0$ \\
$B_1$ & $H{=}1,C{=}0$, wrong   & \cellcolor{gray!22}$+0.5$ & $-0.2$ & $-1.0$ \\
$B_2$ & $H{=}1$, timeout       & \cellcolor{gray!22}$+0.5$ & $-0.1$ & $0$ \\
$C$   & $H{=}0,C{=}0$          & -- & \textsc{Base} & \textsc{Base} \\
$D$   & $H{=}0,C{=}1,\Retr{=}1$ & -- & \cellcolor{gray!22}$-0.2$ & \cellcolor{gray!22}$-0.1$ \\
$D$   & $H{=}0,C{=}1,\Retr{=}0$ & \multicolumn{3}{c}{\textsc{Mask}} \\
\bottomrule
\end{tabular}
\end{table}

\subsection{Bounded, sign-preserving residual}
\label{sec:residual}

For token $t$ of rollout $r_i$, \method uses
\begin{equation}
\label{eq:final}
A_{i,t}^{\mathrm{token}}=\hat A_i+e_{i,t}\,\eta\,|\hat A_i|\,P_{T_i,S(t)}\,\lambda_{T_i,S(t)}\,c_{i,t},
\end{equation}
where $c_{i,t}\in[0,c_{\max}]$ is the token salience of \S\ref{sec:salience}.
The gate $e_{i,t}\in\{0,1\}$ is open only if the group has non-zero reward variance, $T_i\neq C$, the token is not one of eight structural-tag literals (e.g., \texttt{<tool\_call>}, \texttt{</answer>}), and $P_{T_i,S(t)}\lambda_{T_i,S(t)}\neq0$.
$A_{i,t}^{\mathrm{token}}$ replaces $\hat A_i$ in the clipped surrogate of GRPO, or inside the sequence-level objective of GSPO.

\begin{proposition}[Sign preservation]
\label{prop:sign}
Let $|P|\le1$, $0\le\lambda\le\lambda_{\max}$, $0\le c\le c_{\max}$, and $\kappa=\eta\lambda_{\max}c_{\max}<1$. Then $A^{\mathrm{token}}_{i,t}=\hat A_i(1+\epsilon_{i,t})$ with $|\epsilon_{i,t}|\le\kappa$. Hence (i)~every token keeps the sign of $\hat A_i$; (ii)~every token stays in the same branch of the PPO clipping function, so the \method gradient is the outcome gradient reweighted token-wise by factors in $[1-\kappa,1+\kappa]$; and (iii)~groups with $\hat A_i\equiv0$ receive no signal.
\end{proposition}

We use $\eta=0.2$, $\lambda_{\max}=1$, and $c_{\max}=0.5$, i.e., $\kappa=0.1$: \method changes the relative strength (by at most 10\%), not the sign, of outcome-derived token advantages; a positive residual can attenuate negative outcome credit and a negative residual positive outcome credit.
An unbounded variant collapses, with actor entropy rising from 0.67 to 6.6 within 145 steps (Appendix~\ref{app:calibration}).

\paragraph{Why a bounded residual matters.}
By Proposition~\ref{prop:gap}, the net outcome advantage of hit rollouts scales with $\Delta$, whereas the residual that \method adds to their \valids{} tokens is non-negative ($P{=}{+}1$ in every hit type), so its aggregate does not vanish with $\Delta$.
Where the grounding gap is small, this residual can counteract cancellation in the aggregate scalar credit assigned to evidence acquisition, even though it never exceeds 10\% of any single token's credit.
Symmetrically, the negative residual on type-$D$ tokens attenuates, without reversing, the positive outcome credit that unsupported correctness receives ($R_i=1>\mu_{\mathcal G}$).

\subsection{Margin--entropy token salience}
\label{sec:salience}

Within a segment, the residual is distributed by a salience score computed from the old policy.
Let $\ell_t=\log\pi_{\theta_{\mathrm{old}}}(y_t\mid s_t)$ for the sampled token $y_t$, let $\mathcal{C}_t$ be its $K{=}3$ most likely competitors, and let $q$ renormalise $\pi_{\theta_{\mathrm{old}}}$ over its top-$M$ tokens $V^M_t$ ($M{=}32$).
The sampled-token margin and the normalised entropy are
\begin{equation}
\begin{aligned}
\delta_t &= \ell_t-\log\tfrac{1}{K}\textstyle\sum_{v\in\mathcal{C}_t}\pi_{\theta_{\mathrm{old}}}(v\mid s_t),\\
\hat H_t &= -\textstyle\sum_{v\in V^M_t}q_v\log q_v\,/\log M.
\end{aligned}
\end{equation}
With the segment-centred margin $\bar\delta_t=\delta_t-\mathrm{median}_{j\in S}\,\delta_j$ and $d^{\pm}_t=\sigma(\mp\bar\delta_t/\tau)$,
\begin{equation}
\label{eq:salience}
\begin{aligned}
c^{+}_t &= \alpha d^{+}_t+\beta\hat H_t+\gamma d^{+}_t\hat H_t,\\
c^{-}_t &= \alpha d^{-}_t+\beta(1-\hat H_t)+\gamma d^{-}_t(1-\hat H_t),
\end{aligned}
\end{equation}
clamped at $c_{\max}$, with $(\alpha,\beta,\gamma)=(0.6,0.3,0.1)$ and $\tau=0.5$.
Cells with $P{=}{+}1$ use $c^+$, which favours \emph{uncertain but useful} choices (low margin, high entropy, e.g., a pivotal query term); cells with $P{=}{-}1$ use $c^-$, which favours \emph{confident but harmful} choices (high margin, low entropy, e.g., a confidently wrong year).
Since $c^-$ follows from $c^+$ by $\bar\delta\mapsto-\bar\delta$ and $\hat H\mapsto1-\hat H$, positive and negative cells with equal $\lambda$ receive equal budgets.
Salience only redistributes a residual whose sign and bound are already fixed; results are insensitive to $K$ and $M$ (Appendix~\ref{app:topk-topm-robustness}).

\section{Experiments}
\label{sec:exp}

\subsection{Setup}
\label{sec:exp-setup}

\begin{table*}[t]
  \centering
  \small
  \setlength{\tabcolsep}{3.6pt}
  \renewcommand{\arraystretch}{1.12}
  \caption{\textbf{Main results (EM, \%) on seven QA benchmarks.}
  $^\dagger$/$^\star$: in-domain/out-of-domain. Best results in \textbf{bold}. \emph{Initial checkpoint + search} (before task RL) and Search-R1 are run in the same setting as \method; \method rows report the mean and $1\sigma$ over 3 training seeds.}
  \label{tab:main-em}
  \resizebox{\textwidth}{!}{
  \begin{tabular}{@{}lrrrrrrrr@{}}
      \toprule
      \textbf{Methods} & \multicolumn{3}{c}{\textbf{General QA}} & \multicolumn{4}{c}{\textbf{Multi-Hop QA}} & \textbf{Avg.} \\
      \cmidrule(lr){2-4}\cmidrule(lr){5-8}
       & \textbf{NQ$^\dagger$} & \textbf{TriviaQA$^\star$} & \textbf{PopQA$^\star$} & \textbf{HotpotQA$^\dagger$} & \textbf{2Wiki$^\star$} & \textbf{MuSiQue$^\star$} & \textbf{Bamboogle$^\star$} & \\
      \midrule
      \multicolumn{9}{c}{\textbf{Qwen2.5-3B-Instruct}} \\
      \midrule
      Initial checkpoint + search & 16.15 & 25.81 & 16.54 & 12.37 & 10.66 & 4.18 & 12.00 & 13.96 \\
      IRCoT & 11.1 & 31.2 & 20.0 & 16.4 & 17.1 & 6.7 & 24.0 & 18.1 \\
      Rejection Sampling & 29.4 & 48.8 & 33.2 & 24.0 & 23.3 & 5.9 & 21.0 & 26.5 \\
      ZeroSearch & 41.4 & 57.4 & \textbf{44.8} & 27.4 & 30.0 & 9.8 & 11.1 & 31.7 \\
      DeSA & 37.5 & 57.5 & 39.7 & 35.2 & 36.3 & 13.4 & 34.7 & 36.3 \\
      Search-R1 & 47.2 & 62.7 & 40.3 & 36.4 & 34.9 & 12.1 & 28.0 & 37.4 \\
      GiGPO & 42.0 & 59.5 & 42.4 & 36.9 & 37.0 & 12.6 & \textbf{64.1} & 42.1 \\
      \rowcolor{blue!10} \method & \textbf{48.06}\textsubscript{\textpm0.76} & \textbf{63.68}\textsubscript{\textpm0.33} & 42.94\textsubscript{\textpm0.28} & \textbf{44.65}\textsubscript{\textpm0.47} & \textbf{44.12}\textsubscript{\textpm0.57} & \textbf{18.74}\textsubscript{\textpm0.38} & 39.20\textsubscript{\textpm3.72} & \textbf{43.06}\textsubscript{\textpm0.56} \\
      \midrule
      \multicolumn{9}{c}{\textbf{Qwen3-30B-A3B-Instruct}} \\
      \midrule
      Initial checkpoint + search & 30.44 & 55.57 & 34.55 & 36.25 & 36.55 & 14.48 & 42.40 & 35.75 \\
      Search-R1 & 49.61 & 69.04 & 46.74 & 45.70 & 46.14 & 19.57 & 47.20 & 46.29 \\
      \rowcolor{blue!10} \method & \textbf{54.71}\textsubscript{\textpm0.52} & \textbf{71.83}\textsubscript{\textpm0.16} & \textbf{50.12}\textsubscript{\textpm0.22} & \textbf{52.64}\textsubscript{\textpm0.28} & \textbf{54.18}\textsubscript{\textpm0.27} & \textbf{27.84}\textsubscript{\textpm0.55} & \textbf{56.80}\textsubscript{\textpm1.82} & \textbf{52.59}\textsubscript{\textpm0.29} \\
      \bottomrule
  \end{tabular}}
\end{table*}

\begin{table*}[t]
\caption{\textbf{Aggregate grounding metrics (\%, averaged over the seven benchmarks).}
\toolhit{} $=P(H{=}1)$; \hitacc{} $=P(\tilde C{=}1\mid H{=}1)$ with the lenient correctness $\tilde C$ (\S\ref{sec:exp-setup}); \supcorr{} $=$ \toolhit{}$\times$\hitacc{} estimates the supported-correct rate; \supcorr{}/Acc is its share of the lenient accuracy Acc.}
\label{tab:grounded}
\centering
\footnotesize
\setlength{\tabcolsep}{5pt}
\renewcommand{\arraystretch}{1.12}
\begin{tabular}{@{}llccccc@{}}
\toprule
Backbone & Method & EM & \toolhit & \hitacc & \supcorr & \supcorr{}/Acc \\
\midrule
\multirow{3}{*}{Qwen2.5-3B}
  & Initial checkpoint & 13.96 & 43.00 & 55.27 & 23.77 & 75.0\% \\
  & Search-R1         & 37.36 & 45.79 & 65.98 & 30.21 & 76.9\% \\
  \rowcolor{blue!10}
  & \textbf{+\method} & \textbf{43.06} & \textbf{64.68} & \textbf{66.17} & \textbf{42.80} & \textbf{93.4\%} \\
\midrule
\multirow{3}{*}{Qwen3-30B-A3B}
  & Initial checkpoint & 35.75 & 57.20 & 69.37 & 39.68 & 89.3\% \\
  & Search-R1         & 46.29 & 65.09 & 68.05 & 44.29 & 92.3\% \\
  \rowcolor{blue!10}
  & \textbf{+\method} & \textbf{52.59} & \textbf{71.87} & \textbf{72.60} & \textbf{52.18} & \textbf{94.1\%} \\
\bottomrule
\end{tabular}
\end{table*}

\paragraph{Models and training.}
We train Qwen2.5-3B-Instruct~\citep{qwen2.5} with GRPO and the Mixture-of-Experts Qwen3-30B-A3B-Instruct-2507~\citep{qwen3} with GSPO in verl~\citep{sheng2025hybridflow}, on the NQ+HotpotQA mixture of Search-R1 with its hyperparameters (batch size 512, 5 rollouts per prompt, learning rate $10^{-6}$, clip ratio 0.2).
All runs share the prompts and the retrieval setup (E5~\citep{wang2022e5} over the 2018 Wikipedia dump, top-3 passages, at most 5 turns; temperature 1.0 in training and 0 in evaluation); \method changes only the token-level advantage.
\method uses a KL coefficient of $5{\times}10^{-3}$ instead of $10^{-3}$; Search-R1 trained with the larger coefficient is not better (36.61 vs.\ 37.36 EM; Appendix~\ref{app:searchr1-kl}).
Further details are in Appendix~\ref{app:hparams}.

\paragraph{Benchmarks and metrics.}
Following \citet{jin2025searchr1}, we evaluate on NQ~\citep{kwiatkowski2019nq}, TriviaQA~\citep{joshi2017triviaqa}, PopQA~\citep{mallen2023popqa}, HotpotQA~\citep{yang2018hotpotqa}, 2WikiMultiHopQA~\citep{ho2020wikimultihop}, MuSiQue~\citep{trivedi2022musique}, and Bamboogle~\citep{press2023bamboogle} (51,713 questions).
Besides EM we report the lenient accuracy (Acc, an alias-aware substring match) of the Search-R1 evaluation script, whose correctness $\tilde C_i$ differs from the EM correctness $C_i$ (Eq.~\ref{eq:subout}).
The grounding metrics are \toolhit{} $=P(H{=}1)$, \hitacc{} $=P(\tilde C{=}1\mid H{=}1)$, supported correctness $\supcorr=\toolhit\cdot\hitacc$ (which can exceed EM), and the grounded share \supcorr/Acc, i.e., the fraction of correct answers backed by retrieved evidence.

\paragraph{Baselines.}
\emph{Initial checkpoint + search} (no task RL), IRCoT~\citep{trivedi2023ircot}, Rejection Sampling~\citep{jin2025searchr1}, ZeroSearch~\citep{sun2025zerosearch}, DeSA~\citep{wang2025desa}, GiGPO~\citep{feng2025group}, and Search-R1, which we reproduce in exactly the setting of \method.

\smallskip\noindent
Beyond the main comparison (\S\ref{sec:main-results}), we ablate each component (\S\ref{sec:ablation}), test whether the routing magnitudes are hand-tuned (\S\ref{sec:learned-routing}), audit the grounding signal (\S\ref{sec:signal}), and analyse training dynamics (\S\ref{sec:training-dynamics}).

\subsection{Main results}
\label{sec:main-results}

\begin{table*}[t]
\caption{\textbf{Budget-matched leave-one-out ablation} on Qwen2.5-3B (\%, mean over 3 training seeds). \emph{W/o salience} replaces token salience by its per-cell mean; \emph{w/o segment split} merges \valids{} and \redunds{} into one search segment. Best results in \textbf{bold}.}
\label{tab:ablation}
\centering
\small
\setlength{\tabcolsep}{3.4pt}
\renewcommand{\arraystretch}{1.12}
\resizebox{\textwidth}{!}{%
\begin{tabular}{@{}lcccccccccccc@{}}
\toprule
 & \multicolumn{8}{c}{\textbf{EM}} & \multicolumn{4}{c}{\textbf{Grounding}} \\
\cmidrule(lr){2-9}\cmidrule(l){10-13}
\textbf{Variant} & NQ & TriviaQA & PopQA & HotpotQA & 2Wiki & MuSiQue & Bamboogle & Avg. & \toolhit & \hitacc & \supcorr & \supcorr/Acc \\
\midrule
Search-R1 & 47.17 & 62.72 & 40.25 & 36.37 & 34.94 & 12.08 & 28.00 & 37.36 & 45.79 & 65.98 & 30.21 & 76.9\% \\
\quad w/o salience & 45.93 & 62.46 & \textbf{43.14} & 41.98 & 40.89 & 16.52 & \textbf{43.13} & 42.01 & 57.85 & \textbf{66.89} & 38.70 & 87.6\% \\
\quad w/o segment split & 44.86 & 61.38 & 42.54 & 40.21 & 41.52 & 14.14 & 37.51 & 40.31 & 53.81 & 64.50 & 34.71 & 82.0\% \\
\rowcolor{blue!10}
\method (full) & \textbf{48.06} & \textbf{63.68} & 42.94 & \textbf{44.65} & \textbf{44.12} & \textbf{18.74} & 39.20 & \textbf{43.06} & \textbf{64.68} & 66.17 & \textbf{42.80} & \textbf{93.4\%} \\
\bottomrule
\end{tabular}}
\end{table*}

\begin{figure*}[t]
\centering
\includegraphics[width=\textwidth]{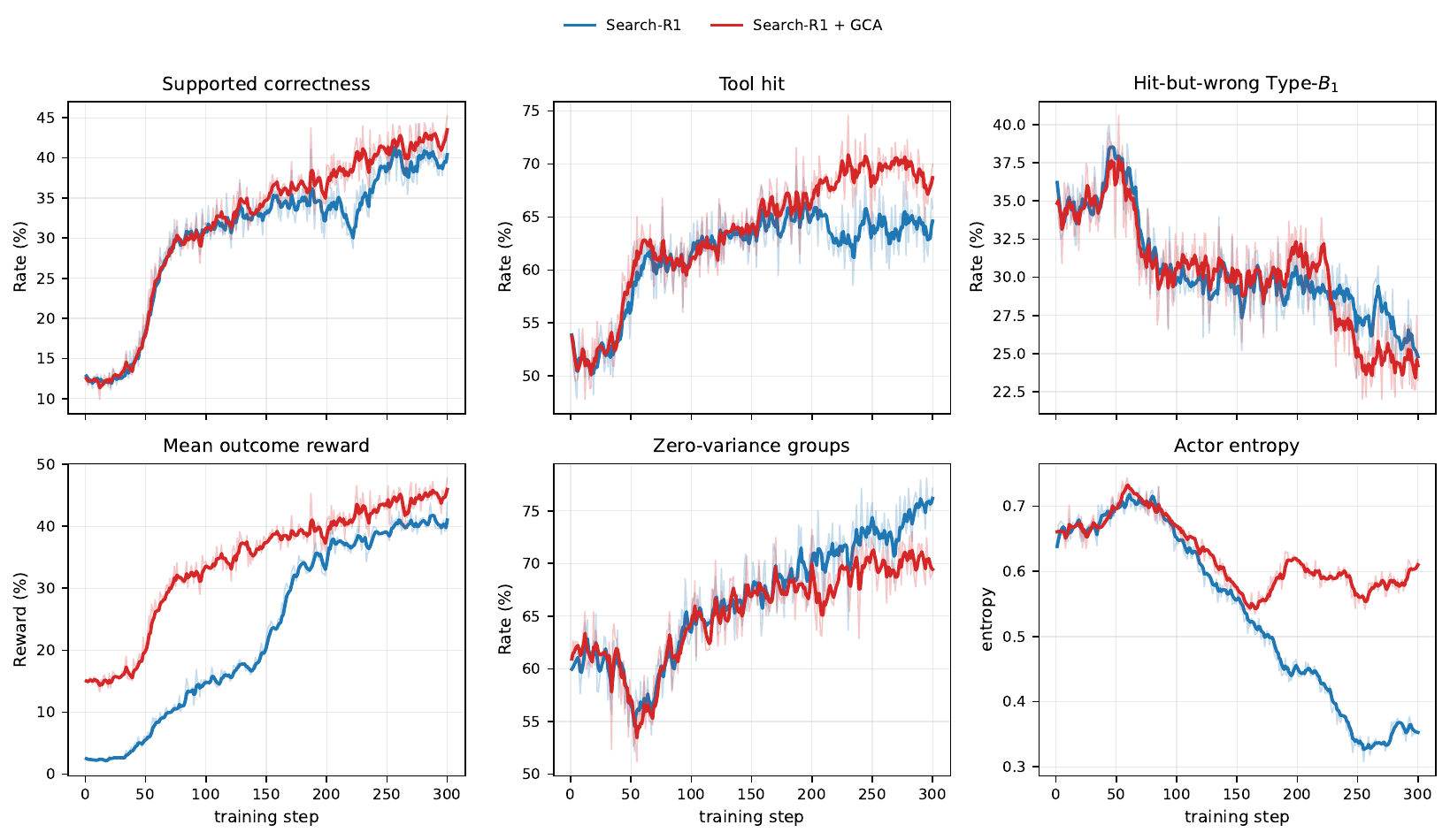}
\caption{\textbf{Training dynamics on Qwen2.5-3B} (\S\ref{sec:training-dynamics}). Faint lines show per-step values and solid lines an exponential moving average. Top: grounded behaviour (\supcorr{}, \toolhit{}, and the hit-but-wrong share of all rollouts, type $B_1$); bottom: optimisation diagnostics (mean outcome reward, zero-variance group rate, and actor entropy).}
\label{fig:training-dynamics}
\end{figure*}

Table~\ref{tab:main-em} shows that \method improves average EM over Search-R1 by 5.70 points on Qwen2.5-3B and 6.30 on Qwen3-30B-A3B under identical settings.
Gains are largest on multi-hop benchmarks, where Search-R1's hit accuracy is lowest (45--65\% vs.\ 67--88\% on single-hop sets), i.e., where hit rollouts carry the most split credit (observation~(ii) in \S\ref{sec:gap}): Bamboogle $+11.2$, 2WikiMultiHopQA $+9.2$, HotpotQA $+8.3$, and MuSiQue $+6.7$ on 3B, and $+9.6$, $+8.0$, $+6.9$, and $+8.3$ on 30B.
GiGPO is the strongest prior method; \method is 1.0 point higher on average but trails it on Bamboogle (39.2 vs.\ 64.1; 125 questions), and since GiGPO groups credit by state and \method by function, the two could be combined.

Grounding improves more than accuracy (Table~\ref{tab:grounded}).
On 3B, \toolhit{} rises from 45.8 to 64.7, \supcorr{} from 30.2 to 42.8, and the grounded share from 76.9\% to 93.4\%; on 30B the three rise from 65.1 to 71.9, 44.3 to 52.2, and 92.3\% to 94.1\%.
The largest per-benchmark gains occur exactly where Search-R1 lost grounding in Figure~\ref{fig:motivation}a: the grounded share rises from 42.9\% to 96.0\% on Bamboogle and from 71.8\% to 98.4\% on MuSiQue (Appendix~\ref{app:detail}).
On 3B, the gain in supported correctness is accompanied primarily by increased retrieval coverage: average \hitacc{} stays nearly unchanged (65.98 vs.\ 66.17), and its changes are heterogeneous across benchmarks (Appendix~\ref{app:detail}).

\subsection{Ablation study}
\label{sec:ablation}

The three levels of \method are nested: routes are indexed by the diagnosed type, and salience redistributes a routed residual, so a variant with salience alone has no defined sign, and removing the diagnosis leaves no route at all ($\eta{=}0$ recovers Search-R1).
We therefore run budget-matched leave-one-out ablations (Table~\ref{tab:ablation}).
\emph{W/o salience} replaces $c_{i,t}$ by its per-cell mean measured on a frozen \method rollout sample, which keeps each cell's residual budget but removes within-cell concentration.
\emph{W/o segment split} merges \valids{} and \redunds{} into one search segment whose signed coefficient preserves their token-weighted residual.

Merging the search segments costs 2.75 EM and 10.9 \toolhit{} points and lowers all seven benchmarks, most on the multi-hop HotpotQA ($-4.4$) and MuSiQue ($-4.6$).
Without the first-hit split, the residual can no longer separate pre-hit search from potentially redundant later search, and the grounded share falls from 93.4\% to 82.0\%.
Removing salience keeps 81.6\% of the EM gain (4.65 of 5.70 points) but loses 6.8 points of \toolhit{} and 5.8 points of grounded share.
It hurts most on NQ, HotpotQA, 2WikiMultiHopQA, and MuSiQue ($-2.1$ to $-3.2$ EM), whereas on Bamboogle, the smallest benchmark (125 questions; $\pm3.7$ across seeds for \method), the constant variant scores higher.
Among the tested ablations, removing the first-hit split causes the largest drop, supporting its empirical importance for separating acquisition-related from potentially redundant search.

\subsection{Learned routing and sensitivity}
\label{sec:learned-routing}

\begin{table}[t]
\caption{\textbf{Fixed vs.\ learned routing magnitudes} on Qwen2.5-3B (average EM, \%, 3 seeds). The learned router keeps the fixed polarities and the residual budget (Appendix~\ref{app:extra-ablation}).}
\label{tab:learned}
\centering
\small
\setlength{\tabcolsep}{4pt}
\begin{tabular}{@{}lccc@{}}
\toprule
Routing & In-domain & Out-of-domain & Overall \\
\midrule
Search-R1 & 41.77 & 35.60 & 37.36 \\
Learned magnitudes & \textbf{46.41} & 41.53 & 42.92 \\
\rowcolor{blue!10}
Fixed (Table~\ref{tab:routing}) & 46.36 & \textbf{41.74} & \textbf{43.06} \\
\bottomrule
\end{tabular}
\end{table}

\begin{table}[t]
\caption{\textbf{Sensitivity} of average EM (\%) to the $\lambda$ scale $s$, the residual scale $\eta$, and the salience cap $c_{\max}$. $\kappa=\eta\lambda_{\max}c_{\max}$ is the worst-case relative residual (Proposition~\ref{prop:sign}); shaded: default; --: not run. Search-R1 scores 37.36 (3B) and 46.29 (30B); \toolhit, \hitacc, and \supcorr{} are in Appendix~\ref{app:sensitivity}.}
\label{tab:sensitivity}
\centering
\small
\setlength{\tabcolsep}{3.6pt}
\begin{tabular}{@{}lccccc@{}}
\toprule
\textit{$\lambda$ scale $s$} & \textit{0.50} & \textit{0.75} & \cellcolor{blue!10}\textit{1.00} & \textit{1.25} & \textit{1.50} \\
$\kappa$ & 0.05  & 0.075 & \cellcolor{blue!10}0.10  & 0.125 & 0.15 \\
EM (3B)  & 41.23 & 42.76 & \cellcolor{blue!10}43.06 & 41.57 & 36.92 \\
EM (30B) & --    & 48.01 & \cellcolor{blue!10}52.59 & 46.77 & -- \\
\midrule
\textit{Residual scale $\eta$} & \textit{0.05} & \textit{0.10} & \cellcolor{blue!10}\textit{0.20} & \textit{0.50} & \textit{1.00} \\
$\kappa$ & 0.025 & 0.05  & \cellcolor{blue!10}0.10  & 0.25  & 0.50 \\
EM (3B)  & 41.98 & 43.72 & \cellcolor{blue!10}43.06 & 40.80 & 37.89 \\
EM (30B) & --    & 51.91 & \cellcolor{blue!10}52.59 & 49.93 & -- \\
\midrule
\textit{Salience cap $c_{\max}$} & \textit{0.20} & \textit{0.30} & \cellcolor{blue!10}\textit{0.50} & \textit{1.00} & \\
$\kappa$ & 0.04  & 0.06  & \cellcolor{blue!10}0.10  & 0.20  & \\
EM (3B)  & 40.83 & 43.07 & \cellcolor{blue!10}43.06 & 42.42 & \\
EM (30B) & --    & 52.00 & \cellcolor{blue!10}52.59 & 50.17 & \\
\bottomrule
\end{tabular}
\end{table}

The polarities are fixed by the rule of \S\ref{sec:routing}; the remaining degrees of freedom are the magnitudes $\lambda$ and the bound $\kappa$.

\paragraph{Learned magnitudes.}
We keep $P$ fixed and learn non-negative magnitudes under the same type-wise budget and bound on a stratified 1\% calibration split (Appendix~\ref{app:extra-ablation}).
The learned router matches the fixed table in-domain and trails it slightly out-of-domain (Table~\ref{tab:learned}): the magnitudes can be learned, but learning brings no gain, so the fixed table is not a tuned optimum.

\paragraph{Sensitivity.}
On 3B, EM stays between 40.8 and 43.7 across a 2.5$\times$ range of $s$, a 4$\times$ range of $\eta$, and a 5$\times$ range of $c_{\max}$ (Table~\ref{tab:sensitivity}).
It falls to the level of Search-R1 only when $s$ or $\eta$ is scaled up until the worst-case residual reaches 15--50\% of $|\hat A_i|$, far beyond the 10\% budget we use.
All tested caps remain above Search-R1.
The 30B model is more sensitive to the $\lambda$ scale (46.8--52.6 EM) but stays above Search-R1 (46.29) in every configuration.

\subsection{Reliability of the grounding signal}
\label{sec:signal}

$H$ is a lexical answer-coverage signal computed on the tool responses the policy actually received (top-3 passages, truncated to 1,024 tokens).
We audited all 169,615 training questions (Appendix~\ref{app:alias-audit}): yes/no answers account for 3.23\% (6.06\% of HotpotQA), multiple aliases for 4.57\% (alternatives, matched disjunctively), and formatting artefacts for 0.97\%.
As a recall test, applying the matcher to the annotated supporting documents of the 90,447 HotpotQA questions misses the alias in 3.61\% of cases, and in only 0.040\% once yes/no answers are excluded, all of them formatting artefacts.
$H$ is therefore a near-perfect recall proxy for genuine evidence except for yes/no verdicts, which are inferred rather than copied.
Retraining without yes/no questions changes in-domain EM by about $+0.7$ (single seed, 100 steps): this noise is mildly harmful but does not drive our gains.
Moreover, by Proposition~\ref{prop:sign} a misdiagnosed rollout changes its tokens' credit by at most 10\% and never flips its sign.

\subsection{Training dynamics}
\label{sec:training-dynamics}

Figure~\ref{fig:training-dynamics} shows how grounded behaviour emerges.
\method raises \supcorr{} and retrieval coverage without a corresponding increase in the overall hit-but-wrong rate in the reported trace.
It also reaches a higher mean outcome reward, leaves fewer groups with zero reward variance (70\% vs.\ 75\% at the end of training), so that more groups carry a learning signal, and keeps actor entropy higher than Search-R1 (0.59 vs.\ 0.36) without the entropy explosion of oversized residuals.
General tool use is preserved: on BFCL v3 the 30B model improves from 68.62 to 69.14 overall, mainly on multi-turn calls (41.12 to 44.12; Appendix~\ref{app:bfcl}).
Appendix~\ref{app:rollout-examples} shows a representative trajectory of each type.

\section{Conclusion}

Under outcome-reward RL, the aggregate advantage favours retrieval-hit rollouts only through the within-group grounding gap, consistent with search agents trained this way gaining accuracy without gaining grounding.
\method adds a bounded, sign-preserving residual whose polarity follows the verifiable sub-outcome of each segment and concentrates it on decision-critical tokens.
Across seven QA benchmarks and two backbones it improves accuracy and, more strongly, the grounding of correct answers, without extra models, rollouts, or training stages.
We expect the same recipe, namely identifying the sub-outcome each segment controls and routing a bounded residual by its polarity, to transfer to agents whose intermediate outcomes are verifiable by other means, such as evidence-entailment judges or executable tests.

\section*{Limitations}

\paragraph{Training-time supervision.}
$H$, $C$, and $\Retr$ are computed from gold answer aliases with a lexical matcher.
$H$ measures answer coverage rather than semantic support: it misses yes/no answers and some comparison answers, which partly explains the low grounded share of all methods on 2WikiMultiHopQA, and it may fire on passages that mention the answer without supporting it.
Templated oracle queries can miss retrievable answers; we mitigate this by setting $\Retr(x){=}1$ whenever a rollout of the group hits.
Gold answers are not needed at inference time, but applying \method where no gold answer exists would require replacing these signals with, e.g., evidence-entailment judges or executable checks, which we have not evaluated.

\paragraph{Scope of the evaluation.}
We study two Qwen-family backbones and a single retrieval environment (E5 over the 2018 Wikipedia dump, top-3 passages, at most five turns); other model families and live web search remain untested.
At 30B we retrain only Search-R1 as a baseline, so that comparison is a controlled transfer test rather than a full benchmark.
Seed variance is reported only for \method.

\paragraph{Design choices.}
One routing table is used for all benchmarks and backbones; learned magnitudes match it (\S\ref{sec:learned-routing}), but we do not ablate every cell individually, and the salience weights $(\alpha,\beta,\gamma)$ are fixed.
The salience score reuses old-policy logits, and we did not measure its overhead.

\bibliography{references}

\appendix

\section{Notation}
\label{app:notation}

\begin{table}[H]
\caption{\textbf{Summary of notation.}}
\label{tab:notation}
\centering
\small
\renewcommand{\arraystretch}{1.12}
\begin{tabular}{@{}c>{\raggedright\arraybackslash}p{0.66\columnwidth}@{}}
\toprule
Symbol & Description \\
\midrule
$r_i$, $\mathcal{G}(x)$ & $i$-th rollout; group of $n$ rollouts for question $x$ \\
$R_i$, $\hat A_i$ & Outcome reward; GRPO/GSPO group-relative advantage \\
$\mathcal{A}(x)$, $m(a,o)$ & Gold alias set; case-insensitive containment test \\
$H_i$, $C_i$ & Retrieval hit (\toolhit); EM answer correctness (Eq.~\ref{eq:subout}) \\
$\tilde C_i$ & Lenient evaluation correctness (alias-aware substring match) used by \hitacc{}, \supcorr{}, and Acc (\S\ref{sec:exp-setup}) \\
$k^{\star}_i$ & First search turn whose tool response contains a gold alias \\
$\Retr(x)$ & Whether the corpus contains the answer (oracle queries) \\
$T_i$ & Rollout type: $A$, $B_1$, $B_2$, $C$, or $D$ \\
$\Delta$ & Within-group grounding gap $\Pr[C\mid H{=}1]-\Pr[C\mid H{=}0]$ \\
$S(t)$ & Functional segment of token $t$: \valids, \redunds, or \answ \\
$P_{T,S}$, $\lambda_{T,S}$ & Routing polarity and magnitude (Table~\ref{tab:routing}) \\
$e_{i,t}$ & Residual eligibility gate \\
$\delta_t$, $\hat H_t$ & Sampled-token margin; top-$M$ normalised entropy \\
$c_{i,t}$ & Token salience ($c^+$ for positive, $c^-$ for negative cells) \\
$\eta$, $c_{\max}$, $\kappa$ & Residual scale (0.2); salience cap (0.5); bound $\eta\lambda_{\max}c_{\max}$ (0.1) \\
$\tau$ & Margin temperature (0.5) \\
$A_{i,t}^{\mathrm{token}}$ & Final token-level advantage (Eq.~\ref{eq:final}) \\
\bottomrule
\end{tabular}
\end{table}

\section{Algorithm summary}
\label{app:algo}

\begin{algorithm}[H]
\caption{One-pass \method for a single question $x$.}
\label{alg:gca}
\begin{algorithmic}[1]
\State Sample group $\mathcal{G}(x)=\{r_1,\ldots,r_n\}$ with $\pi_{\theta_{\mathrm{old}}}$.
\For{each rollout $r_i$}
  \State Compute $H_i$, $C_i$; classify primary type $T_i \in \{A, B, C, D\}$.
  \State For Type~$B$, record whether it is valid-wrong-answer or timeout.
\EndFor
\State Compute oracle-query retrievability $\Retr(x)$; set to $1$ if any $H_i{=}1$.
\State Compute $R_i = R_{\mathrm{SearchR1}}(r_i)$ and $\hat A_i = \mathrm{Adv}_{\mathrm{GRPO/GSPO}}(R_i)$.
\State Segment each assistant token into $\{\valids, \redunds, \answ\}$ via the first-hit index $k^\star$.
\State Mark residual-eligible response tokens by excluding only the eight structural-tag literals in Appendix~\ref{app:routing}.
\For{each rollout $r_i$ with $T_i \neq C$ and $\mathrm{std}(R_{\mathcal{G}})>\epsilon$}
  \State Look up $(P_{T_i,S}, \lambda_{T_i,S})$ from Table~\ref{tab:routing}.
  \For{each residual-eligible token $t$}
    \State Compute positive salience $c_{i,t}^+$ if $P_{T_i,S(t)}{>}0$, or negative salience $c_{i,t}^-$ if $P_{T_i,S(t)}{<}0$.
    \State $A_{i,t}^{\mathrm{token}} \gets \hat A_i + \eta\,|\hat A_i|\,P_{T_i,S}\,\lambda_{T_i,S}\,c_{i,t}$.
  \EndFor
  \State Set $A_{i,t}^{\mathrm{token}} \gets \hat A_i$ for excluded structural-tag literals and unrouted cells.
\EndFor
\State For remaining rollouts (Type~$C$, homogeneous groups): $A_{i,t}^{\mathrm{token}} \gets \hat A_i$.
\State Take a GRPO/GSPO step with the token-level advantages.
\end{algorithmic}
\end{algorithm}

\section{Proofs}
\label{app:proofs}

\paragraph{Proposition~\ref{prop:gap}.}
Let $n_H=n p_H$ and write $\mu$ for the group mean of $R_i=C_i$. Then
\begin{equation*}
\begin{aligned}
\sum_{i:H_i=1}\hat A_i&=\frac{1}{\sigma}\sum_{i:H_i=1}(C_i-\mu)\\
&=\frac{n_H}{\sigma}\bigl(\Pr[C\mid H{=}1]-\mu\bigr).
\end{aligned}
\end{equation*}
Since $\mu=p_H\Pr[C\mid H{=}1]+(1-p_H)\Pr[C\mid H{=}0]$, we have $\Pr[C\mid H{=}1]-\mu=(1-p_H)\Delta$, which gives the claim.
The second equality follows from $\sum_i\hat A_i=0$.
For a general reward $R_i$ (e.g., with a format term), the same computation gives $\sum_{i:H_i=1}\hat A_i=n\,\mathrm{Cov}(H,R)/\sigma$, where $\mathrm{Cov}$ is the within-group covariance; for $R=C$, $\mathrm{Cov}(H,R)=p_H(1-p_H)\Delta$.
Implementations that add a small constant to $\sigma$ or use the sample standard deviation change only the positive prefactor. \hfill$\square$

\paragraph{Proposition~\ref{prop:sign}.}
If $e_{i,t}=1$, then $A^{\mathrm{token}}_{i,t}=\hat A_i+\eta|\hat A_i|P\lambda c=\hat A_i\bigl(1+\eta\,\mathrm{sign}(\hat A_i)P\lambda c\bigr)$, so $\epsilon_{i,t}=\eta\,\mathrm{sign}(\hat A_i)P\lambda c$ and $|\epsilon_{i,t}|\le\eta\lambda_{\max}c_{\max}=\kappa$; if $e_{i,t}=0$, $\epsilon_{i,t}=0$.
(i)~Since $\kappa<1$, $1+\epsilon_{i,t}>0$.
(ii)~For a token with ratio $\rho$ and advantage $A$, the clipped surrogate is $A\min(\rho,1+\epsilon_{\mathrm{clip}})$ if $A>0$ and $A\max(\rho,1-\epsilon_{\mathrm{clip}})$ if $A<0$; the active branch depends only on $\mathrm{sign}(A)$ and $\rho$, which \method leaves unchanged, so each token's gradient is multiplied by $1+\epsilon_{i,t}$.
For GSPO the ratio and clipping are sequence-level and therefore identical for all tokens of a rollout.
(iii)~If $\hat A_i=0$, then $A^{\mathrm{token}}_{i,t}=0$. \hfill$\square$

%

\section{Polarity and Importance Coefficient Analysis}
\label{app:routing}


\subsection{EDA observations on Search-R1 checkpoints}
\label{app:eda-observations}

We profile a Search-R1-trained Qwen2.5-3B run at four checkpoints
(\textsc{base}, \textsc{step300}, \textsc{step600}, \textsc{step900}) and
record, per rollout type $T$ and per segment
$S\!\in\!\{\valids,\redunds,\answ\}$, the number of trainable tokens
$n_{T,S}$.
``Trainable'' means response tokens minus the eight structural-tag
literals
(\texttt{<answer>}/\texttt{</answer>}, \texttt{<tool\_call>}/\texttt{</tool\_call>},
\texttt{<|im\_start|>}/\texttt{<|im\_end|>}, \texttt{<think>}/\texttt{</think>});

\paragraph{Sampling rule.}
For each
checkpoint we draw $100$ questions per evaluation benchmark
(\textsc{nq}, \textsc{triviaqa}, \textsc{popqa}, \textsc{hotpotqa},
\textsc{2wikimultihopqa}, \textsc{musique}, \textsc{bamboogle}) with a
deterministic seed, run pass@$5$ rollouts at $\mathrm{temperature}\!=\!1.0$
with up to $5$ tool turns and a $768$-token per-turn cap, and serve the
checkpoint via vLLM with $\mathrm{top\!-\!}K\!=\!64$ logprobs per position.
The resulting per-token records define the rollout-type and segment
labels used throughout this appendix.

\begin{table*}[t]
\caption{\textbf{Per-cell trainable token counts $n_{T,S}$ and within-type
share at four checkpoints of the Search-R1 Qwen2.5-3B run.}
``Trainable'' = response tokens minus the eight structural-tag literals.
Percentages are within-row ratios $n_{T,S}/\sum_{S'}n_{T,S'}$ for that
rollout type at that checkpoint; type $C$ has no segment partition.}
\label{tab:app-token-counts}
\centering
\footnotesize
\setlength{\tabcolsep}{4pt}
\begin{tabular}{@{}lrrrr@{}}
\toprule
Cell $(T,S)$ & Base & Step\,300 & Step\,600 & Step\,900 \\
\midrule
$A\!\cdot\!\valids$    & 114{,}426 (30.68\%) & 74{,}815 (36.44\%)  & 100{,}449 (25.62\%) & 102{,}219 (30.07\%) \\
$A\!\cdot\!\redunds$   & 100{,}010 (26.81\%) & 54{,}214 (26.41\%)  & 116{,}053 (29.60\%) & 111{,}889 (32.91\%) \\
$A\!\cdot\!\answ$      & 158{,}538 (42.51\%) & 76{,}254 (37.15\%)  & 175{,}504 (44.77\%) & 125{,}836 (37.02\%) \\
$B_1\!\cdot\!\valids$  & 113{,}462 (27.97\%) & 49{,}766 (29.26\%)  & 69{,}547 (23.23\%)  & 61{,}024 (24.59\%)  \\
$B_1\!\cdot\!\redunds$ & 102{,}110 (25.18\%) & 57{,}275 (33.67\%)  & 91{,}709 (30.63\%)  & 97{,}712 (39.37\%)  \\
$B_1\!\cdot\!\answ$    & 190{,}023 (46.85\%) & 63{,}066 (37.07\%)  & 138{,}163 (46.14\%) & 89{,}439 (36.04\%)  \\
$B_2\!\cdot\!\valids$  & 4{,}297 (41.71\%)   & 510 (80.19\%)       & 1{,}893 (79.57\%)   & 2{,}204 (28.72\%)   \\
$B_2\!\cdot\!\redunds$ & 6{,}006 (58.29\%)   & 126 (19.81\%)       & 486 (20.43\%)       & 5{,}471 (71.28\%)   \\
$D\!\cdot\!\redunds$   & 24{,}239 (65.26\%)  & 20{,}366 (63.97\%)  & 19{,}671 (57.45\%)  & 25{,}893 (63.58\%)  \\
$D\!\cdot\!\answ$      & 12{,}903 (34.74\%)  & 11{,}471 (36.03\%)  & 14{,}569 (42.55\%)  & 14{,}831 (36.42\%)  \\
$C$ (no residual)      & 707{,}865 & 348{,}861 & 525{,}512 & 443{,}078 \\
\bottomrule
\end{tabular}
\end{table*}

\begin{figure*}[t]
\centering
\includegraphics[width=0.98\linewidth]{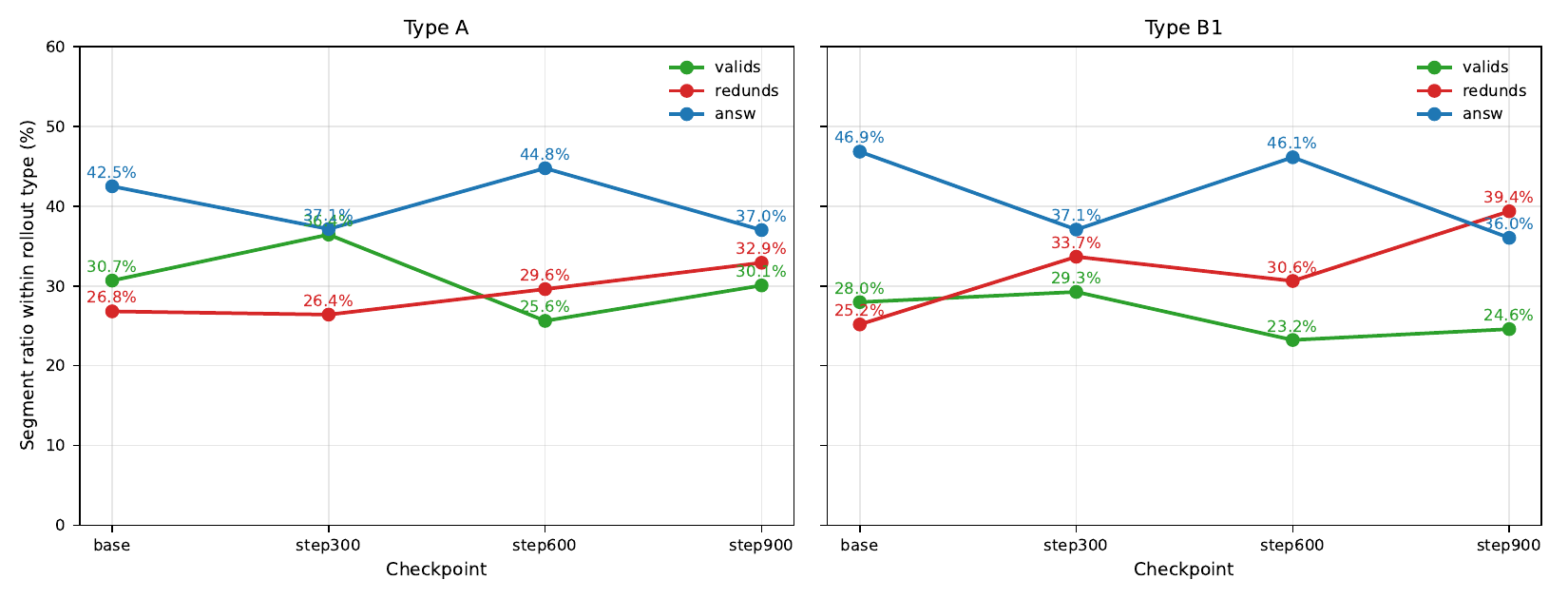}
\caption{Within-row segment ratios for type $A$ (left) and type $B_1$
(right) across the four checkpoints. Percentages denote
$n_{T,S}/\sum_{S'}n_{T,S'}$.}
\label{fig:app-r1-A-B1}
\end{figure*}

\begin{figure*}[t]
\centering
\includegraphics[width=0.98\linewidth]{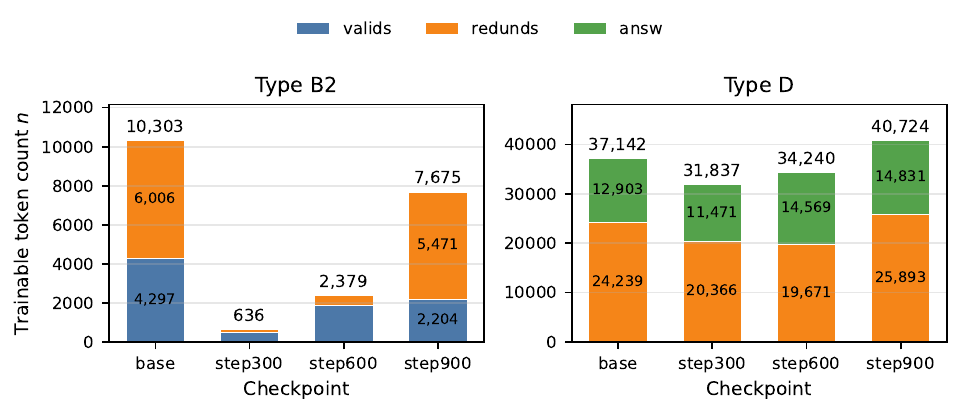}
\caption{Absolute trainable-token counts for type $B_2$ (left) and type
$D$ (right) across the four checkpoints, with stacked segment composition.}
\label{fig:app-r1-B2-D}
\end{figure*}

\paragraph{Findings.}
Four trends emerge from
Figs.~\ref{fig:app-r1-A-B1}--\ref{fig:app-r1-B2-D} and
Table~\ref{tab:app-token-counts}.
(i)~\emph{Even on supported-correct rollouts, post-hit redundant search
grows.} The within-$A$ share of $\redunds$ increases overall across
checkpoints, from $26.81\%$ to $32.91\%$, following a small initial
decrease to $26.41\%$, counter to the
naive expectation that successful rollouts should taper off retrieval
once the answer is grounded.
(ii)~\emph{On $\toolhit$-but-wrong rollouts ($B_1$), the trajectory tilts
away from useful search and toward redundant search.}
$B_1\!\cdot\!\redunds$ rises overall
($25.18\%\!\to\!33.67\%\!\to\!30.63\%\!\to\!39.37\%$) while
$B_1\!\cdot\!\valids$ falls overall
($27.97\%\!\to\!29.26\%\!\to\!23.23\%\!\to\!24.59\%$); the $B_1$ rollouts
are spending an increasing fraction of their budget on retrievals that do
not change the answer outcome.
(iii)~\emph{$B_2$ token mass collapses then rebounds.}
$B_2$ total tokens fall from $10{,}303$ (base) to $636$ (step\,300) and
then climb back to $7{,}675$ (step\,900); the step\,900 $\redunds$ count
($5{,}471$) is within $9\%$ of the base count ($6{,}006$), signalling a
relapse of the post-hit redundant-search pattern in timeout rollouts.
(iv)~\emph{Type $D$ tokens grow in absolute terms.}
$D\!\cdot\!\redunds$ at step\,900 ($25{,}893$) exceeds the base
($24{,}239$), and $D\!\cdot\!\answ$ rises by $\sim\!15\%$ over the same
window; under outcome-only RL, this is the trajectory through which
\unsupcorr{} accrues positive advantage despite a failed retrieval.
These trends motivate a residual that (a) suppresses $\redunds$ on
$A$, $B_1$, $B_2$ and $D$, (b) protects $\valids$ on $A$ and $B_1$, and
(c) softly suppresses $D\!\cdot\!\answ$ when retrievable.

\subsection{Polarity and importance from per-segment diagnosis}
\label{app:per-segment}

The polarity $P_{T,S}$ is set by the proxy sub-outcome rule of
\S\ref{sec:routing}, and the trends in \S\ref{app:eda-observations} indicate
where outcome-only credit is misaligned; the importance ordering follows
two rules:
\begin{itemize}[leftmargin=1.6em]
\item \textbf{Blame the bottleneck} -- the segment most responsible for
the type's outcome receives the largest $|P\lambda|$ in its row.
\item \textbf{Preserve useful sub-behaviour} -- in failed-answer
rollouts, assign a positive residual to valid search.
\end{itemize}

With polarities and the within-row ordering of $\lambda$ fixed by the
two rules, the token-level residual of \method factors as
\begin{equation}
\begin{aligned}
\Delta A_{i,t}^{\mathrm{GCA}}
&=\;
\eta\,|\hat A_i|
\;\cdot\;
P_{T_i,S(t)}
\;\cdot\;
\lambda_{T_i,S(t)}
\;\cdot\;
c_{i,t},\\
&\quad
P\!\in\!\{-1,0,{+}1\},\;
\lambda\!\ge\!0,\;
c_t\!\in\![0,c_{\max}].
\end{aligned}
\end{equation}
The polarity $P_{T,S}$ encodes the design-time direction of the residual
(reinforce vs.\ suppress); the importance $\lambda_{T,S}$ specifies the
relative magnitude across segments; and the salience $c_t$ concentrates
the within-cell residual on decision-critical tokens
(\S\ref{app:salience}).
The composite is bounded by
$\eta\,|\hat A_i|\,\lambda_{\max}\,c_{\max}$.
Table~\ref{tab:app-routing-final} fixes the values used throughout this
work.

\begin{table*}[t]
\caption{\textbf{Routing coefficients $(P_{T,S},\lambda_{T,S})$ used in all
reported experiments.}
$c^{\pm}$ indicates the salience variant applied (\S\ref{app:salience}).
\textit{Base} = original GRPO/GSPO advantage only;
\textit{Mask} = removed from the policy update;
\textit{N/A} = segment absent for this type.
We use $\eta\!=\!0.2$ and $c_{\max}\!=\!0.5$, giving worst-case residual
$\eta\,\lambda_{\max}\,c_{\max}\,|\hat A_i| = 0.1\,|\hat A_i|$.}
\label{tab:app-routing-final}
\centering
\small
\renewcommand{\arraystretch}{1.15}
\begin{tabular}{@{}c l c c c@{}}
\toprule
Type & Condition & \valids & \redunds & \answ \\
\midrule
$A$   & $H{=}1,C{=}1$
      & $P{=}{+}1,\;\lambda{=}0.5,\;c^+$
      & $P{=}{-}1,\;\lambda{=}0.1,\;c^-$
      & $P{=}{+}1,\;\lambda{=}1.0,\;c^+$ \\
$B_1$ & $H{=}1,C{=}0$, valid wrong
      & $P{=}{+}1,\;\lambda{=}0.5,\;c^+$
      & $P{=}{-}1,\;\lambda{=}0.2,\;c^-$
      & $P{=}{-}1,\;\lambda{=}1.0,\;c^-$ \\
$B_2$ & $H{=}1,C{=}0$, timeout
      & $P{=}{+}1,\;\lambda{=}0.5,\;c^+$
      & $P{=}{-}1,\;\lambda{=}0.1,\;c^-$
      & $P{=}0,\;\lambda{=}0.0$ \\
$C$   & $H{=}0,C{=}0$
      & N/A & base & base \\
$D$   & $H{=}0,C{=}1$, Retr$=1$
      & N/A
      & $P{=}{-}1,\;\lambda{=}0.2,\;c^-$
      & $P{=}{-}1,\;\lambda{=}0.1,\;c^-$ \\
$D$   & $H{=}0,C{=}1$, Retr$=0$
      & masked & masked & masked \\
\bottomrule
\end{tabular}
\end{table*}

\subsection{Calibration constraints from observed failure modes}
\label{app:calibration}

The values in Table~\ref{tab:app-routing-final} are constrained by four
failure modes observed in long-horizon training.
We document them as design constraints rather than as a tuning trajectory.

\begin{figure*}[t]
\centering
\includegraphics[width=0.7\linewidth]{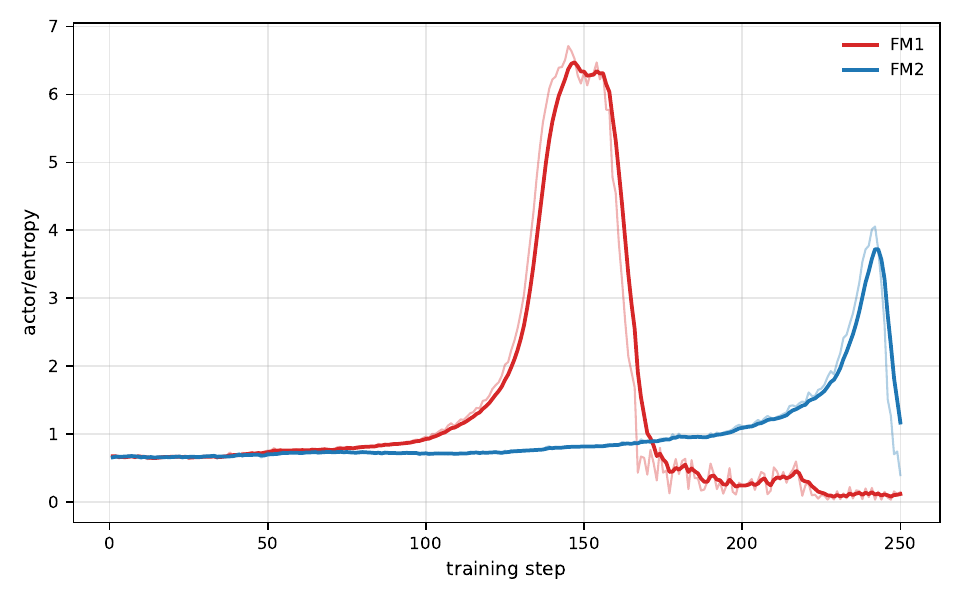}
\caption{Actor entropy traces of the FM1 and FM2 diagnostic runs on
Qwen2.5-3B. FM1 ($\lambda_{B_1,\answ}\!=\!2.0,\,\eta\!=\!0.5$, no $c_t$
cap) reaches its peak around step\,$145$ and collapses thereafter; FM2
keeps the same $\lambda$ but halves $\eta$ to $0.2$, which only delays
the explosion.}
\label{fig:app-fm1-fm2-entropy}
\end{figure*}

\paragraph{FM1: asymmetric format pressure on a high-$\lambda$ negative cell.}
\label{app:fm1}
A negative-residual cell with large $\lambda$ acts on every trainable
token, including format-defining action tokens shared with positive-residual
cells in other rollout types of the same group.
When the per-token negative residual approaches $|\hat A_i|$, probability
mass redistributes off the shared action tokens onto the rest of the
vocabulary; entropy rises steadily until the policy degenerates.
A diagnostic run with $\lambda_{B_1,\answ}\!=\!2.0$, $\eta\!=\!0.5$, no
$c_t$ cap and $\beta_{\mathrm{KL}}\!=\!10^{-3}$ exhibited the canonical
signature -- entropy from $0.67$ to $6.6$ over $145$ steps, then mode
collapse (red curve in Fig.~\ref{fig:app-fm1-fm2-entropy}).
\newline
\textbf{Constraint:} $\eta\,\lambda_{\max}\,c_{\max}\!\le\!0.1$.

\paragraph{FM2: $\lambda$ and $\eta$ are not interchangeable.}
\label{app:fm2}
Halving $\eta$ rescales every cell's residual uniformly but does not
redistribute mass between cells.
A run that retained $\lambda_{B_1,\answ}\!=\!2.0$ and reduced $\eta$ from
$0.5$ to $0.2$ delayed the FM1 entropy explosion -- the $5\!\times\!$
initial-entropy crossing moved from step $\sim\!130$ to step $\sim\!230$
(blue curve in Fig.~\ref{fig:app-fm1-fm2-entropy}) -- but did not prevent
it; the run still collapsed shortly afterwards.
\newline
\textbf{Constraint:} fix $\lambda$ ratios by the per-cell rule first;
choose $\eta$ second, subject to FM1.

\paragraph{FM3: per-token salience extremes dominate without a cap.}
\label{app:fm3}
The salience score $c_t$ is bounded analytically in $[0,1]$ but its
empirical distribution is right-skewed: at the reported checkpoint,
$B_1\!\cdot\!\answ$ has mean $c_t^-\!=\!0.58$ but $49.6\%$ of tokens lie
above $0.5$ and the top $5\%$ carry $8.7\%$ of the cell's uncapped
residual mass.
Without a cap, repeatedly-extreme tokens (BPE wrappers, common prose)
accumulate gradient that drives FM1 even at moderate cell-level $\lambda$.
\newline
\textbf{Constraint:} $c_t\!\le\!c_{\max}\!=\!0.5$; uniform across cells.

\paragraph{FM4: KL anchor must scale with the residual.}
\label{app:fm4}
The KL coefficient $\beta_{\mathrm{KL}}\!=\!10^{-3}$ commonly used for
vanilla GRPO yields a KL-loss to policy-gradient ratio of
$\sim\!10^{-3}$, leaving the residual gradient effectively unanchored.
KL grew $440\!\times\!$ during the FM1 reference run.
\newline
\textbf{Constraint:} $\beta_{\mathrm{KL}}\!\ge\!5\!\times\!10^{-3}$ when
\method{} is enabled, scaling proportionally with the residual budget.

\subsection{Sensitivity analysis}
\label{app:sensitivity}

We report sensitivity along three axes: a multiplicative scale $s$ on the
active $\lambda$ table (Table~\ref{tab:app-lambda-sweep}), the global
residual scale $\eta$ (Table~\ref{tab:app-eta-sweep}), and the per-token
cap $c_{\max}$ (Table~\ref{tab:app-cap-sweep}).
\supcorr{} is the primary diagnostic. Scaling $s$ strengthens every active
residual without changing its polarity, so the $s$ sweep measures
sensitivity to the overall residual budget.

\begin{table}[H]
\caption{Sensitivity to a global $\lambda$ scale $s$ on Qwen-family backbones.
$s\!=\!1$ corresponds to Table~\ref{tab:app-routing-final}.}
\label{tab:app-lambda-sweep}
\centering
\small
\resizebox{\columnwidth}{!}{%
\begin{tabular}{@{}lcccc@{}}
\toprule
$s$ & EM & \toolhit & \hitacc & \supcorr \\
\midrule
\multicolumn{5}{@{}l}{\textit{Qwen2.5-3B}} \\
Search-R1 (no residual) & 37.36 & 45.79 & 65.98 & 30.21 \\
$0.50$ & 41.23 & 61.80 & 65.44 & 40.44 \\
$0.75$ & 42.76 & 62.02 & 66.75 & 41.40 \\
$\mathbf{1.00}$ & \textbf{43.06} & \textbf{64.68} & \textbf{66.17} & \textbf{42.80} \\
$1.25$ & 41.57 & 59.57 & 67.84 & 40.41 \\
$1.50$ & 36.92 & 44.43 & 65.89 & 29.27 \\
\midrule
\multicolumn{5}{@{}l}{\textit{Qwen3-30B-A3B-Instruct}} \\
Search-R1 (no residual) & 46.29 & 65.09 & 68.05 & 44.29 \\
$0.75$ & 48.01 & 69.26 & 68.93 & 47.74 \\
$\mathbf{1.00}$ & \textbf{52.59} & \textbf{71.87} & \textbf{72.60} & \textbf{52.18} \\
$1.25$ & 46.77 & 67.54 & 67.91 & 45.86 \\
\bottomrule
\end{tabular}}
\end{table}

\begin{table}[H]
\caption{Sensitivity to the global residual scale $\eta$ on Qwen-family backbones.}
\label{tab:app-eta-sweep}
\centering
\small
\begin{tabular}{@{}lcccc@{}}
\toprule
$\eta$ & EM & \toolhit & \hitacc & \supcorr \\
\midrule
\multicolumn{5}{@{}l}{\textit{Qwen2.5-3B}} \\
$0.05$ & 41.98 & 63.01 & 65.76 & 41.44 \\
$0.10$ & 43.72 & 63.61 & 65.95 & 41.95 \\
$\mathbf{0.20}$ & \textbf{43.06} & \textbf{64.68} & \textbf{66.17} & \textbf{42.80} \\
$0.50$ & 40.80 & 62.61 & 64.67 & 40.48 \\
$1.00$ & 37.89 & 59.45 & 63.14 & 37.53 \\
\midrule
\multicolumn{5}{@{}l}{\textit{Qwen3-30B-A3B-Instruct}} \\
$0.10$ & 51.91 & 71.64 & 71.19 & 51.00 \\
$\mathbf{0.20}$ & \textbf{52.59} & \textbf{71.87} & \textbf{72.60} & \textbf{52.18} \\
$0.50$ & 49.93 & 70.42 & 70.62 & 49.73 \\
\bottomrule
\end{tabular}
\end{table}

\begin{table}[H]
\caption{Sensitivity to the per-token cap $c_{\max}$ on Qwen-family backbones.}
\label{tab:app-cap-sweep}
\centering
\small
\begin{tabular}{@{}lcccc@{}}
\toprule
$c_{\max}$ & EM & \toolhit & \hitacc & \supcorr \\
\midrule
\multicolumn{5}{@{}l}{\textit{Qwen2.5-3B}} \\
$0.20$ & 40.83 & 64.42 & 62.77 & 40.43 \\
$0.30$ & 43.07 & 64.58 & 65.12 & 42.05 \\
$\mathbf{0.50}$ & \textbf{43.06} & \textbf{64.68} & \textbf{66.17} & \textbf{42.80} \\
$1.00$ & 42.42 & 63.75 & 65.69 & 41.88 \\
\midrule
\multicolumn{5}{@{}l}{\textit{Qwen3-30B-A3B-Instruct}} \\
$0.30$ & 52.00 & 72.96 & 70.70 & 51.58 \\
$\mathbf{0.50}$ & \textbf{52.59} & \textbf{71.87} & \textbf{72.60} & \textbf{52.18} \\
$1.00$ & 50.17 & 70.09 & 70.84 & 49.65 \\
\bottomrule
\end{tabular}
\end{table}

%
%

\section{Token-Level Salience Score Design}
\label{app:salience}


\subsection{Building blocks: sampled-token margin and Top-$M$ entropy}
\label{app:magic-building-blocks}

Let $\ell_t \!=\! \log\pi_{\theta_{\mathrm{old}}}(y_t\!\mid\! s_t)$ denote the
old-policy log-probability of the sampled token $y_t$ at decoding step $t$,
and let $\mathcal{V}$ denote the full vocabulary. We compute two
old-policy diagnostics on every trainable assistant token.

\paragraph{Sampled-token margin $\delta_t$.}
Let $\mathcal{C}_t \!=\! \mathrm{TopK}(\mathcal{V}\setminus\{y_t\})$ collect the
$K$ highest-probability \emph{competing} tokens at step $t$. The competitor
score is the log-mean-exp over $\mathcal{C}_t$,
\begin{equation}
\begin{aligned}
\bar{\ell}^{\mathrm{comp}}_t
&=\log\!\left(\frac{1}{K}\sum_{v\in\mathcal{C}_t}\exp(\log\pi_{\theta_{\mathrm{old}}}(v\!\mid\! s_t))\right),\\
&\qquad K=3,
\end{aligned}
\label{eq:app-comp}
\end{equation}
and the sampled-token margin is the gap between the chosen log-prob and the
competitor average,
\begin{equation}
\delta_t \;=\; \ell_t \;-\; \bar{\ell}^{\mathrm{comp}}_t.
\label{eq:app-margin}
\end{equation}
A small $\delta_t$, whether negative or near zero, indicates that the sampled token
is comparable to or weaker than the top alternatives: under the old policy,
$y_t$ was a \emph{low-confidence} commitment whose informational value, if it
contributed to a useful trajectory, was high relative to its sampling
probability.
A large $\delta_t$ indicates a confident commitment in which $y_t$
dominates the competitors; if it contributed to an undesirable trajectory,
the policy is overcommitted and a stronger negative residual is warranted.

\paragraph{Top-$M$ normalised entropy $\hat H_t$.}
Let $V_t^M \!=\! \mathrm{TopM}(\mathcal{V})$ be the $M$ most likely tokens at
step $t$ and let $q_v\!=\!\pi_{\theta_{\mathrm{old}}}(v|s_t)/\!\!\sum_{u\in V_t^M}\!\!\pi_{\theta_{\mathrm{old}}}(u|s_t)$ be the
re-normalised top-$M$ distribution. We compute
\begin{equation}
\hat H_t \;=\; \frac{-\sum_{v\in V_t^M} q_v\log q_v}{\log M}\;\in[0,1],
\quad M=32.
\label{eq:app-entropy}
\end{equation}
$\hat H_t$ quantifies the spread of probability mass over the local
alternatives. We use top-$M$ rather than full-vocabulary entropy because the
tail of $\pi_{\theta_{\mathrm{old}}}$ contributes negligible mass to the
decoding distribution but inflates the denominator of any normalised
entropy; truncating at $M\!=\!32$ keeps the dynamic range of $\hat H_t$
informative throughout training.

\paragraph{Why both signals are needed.}
The two diagnostics measure related but \emph{not interchangeable} aspects
of local uncertainty.
$\hat H_t$ depends only on the \emph{shape} of $\pi_{\theta_{\mathrm{old}}}$,
not on which token was sampled: it cannot distinguish a low-margin sample
in a flat distribution from a high-margin sample in the same flat
distribution.
$\delta_t$ is sample-conditioned and identifies whether the chosen token is
a winner against its near-neighbours, but it does not measure the breadth of
the alternative set: a margin of $\delta_t \!=\! 5$ nats has very different
implications when the next-best three tokens carry $90\%$ residual mass
versus when they carry $5\%$.
Combining the two signals (\S\ref{app:magic-pos}--\S\ref{app:magic-neg})
recovers the joint information that either signal alone discards.

\subsection{Positive salience $c_t^{+}$ for positive residual segments}
\label{app:magic-pos}

For a token $t$ inside an active positive-residual segment $S$ where
$P_{T,S}\!=\!{+}1$, the salience should be \emph{large} when the token is a
candidate \emph{low-probability, high-value} write -- i.e.\ when the
chosen-token margin is low relative to its segmental neighbourhood and the
surrounding distribution is broad enough that the choice was non-trivial.

Within each active segment $S$ we centre the margin by its segment-local
median to neutralise across-segment scale shift,
\begin{equation}
\bar\delta_t \;=\; \delta_t \;-\; \mathrm{median}_{j\in S}(\delta_j),
\label{eq:app-centered-margin}
\end{equation}
and apply a temperature-scaled, polarity-flipped sigmoid:
\begin{equation}
d_t^{+} \;=\; \sigma\!\left(-\bar\delta_t/\tau_\delta\right),\quad
\tau_\delta = 0.5,\quad d_t^{+}\!\in\!(0,1).
\label{eq:app-d-pos}
\end{equation}
$d_t^{+}$ saturates near $1$ when $\bar\delta_t \!\ll\! 0$
(below-segment-median margin, i.e.\ the token was a relatively
\emph{uncertain} pick within the segment) and decays smoothly toward $0$ for
above-median margins. The temperature $\tau_\delta\!=\!0.5$ produces a
$\pm 1$-nat half-saturation window.

The positive salience composes $d_t^{+}$ with the entropy term:
\begin{equation}
\boxed{
\begin{aligned}
c_t^{+} \;=\; I_t^{+} &\;=\; \alpha\, d_t^{+}\;+\;\beta\,\hat H_t\;+\;\gamma\, d_t^{+}\,\hat H_t,\\
(\alpha,\beta,\gamma) &\;=\; (0.6,\,0.3,\,0.1).
\end{aligned}
}
\label{eq:app-magic-pos}
\end{equation}
Reading the three terms:
\begin{enumerate}[leftmargin=1.6em,topsep=2pt,itemsep=2pt]
\item $\alpha\, d_t^{+}$ is the \emph{margin term}: it rewards tokens that
were low-confidence picks within their segment. Carrying the largest
weight ($\alpha\!=\!0.6$) reflects the empirical observation
(\S\ref{app:why-margin}) that margin remains a stable per-cell
discriminator throughout training: the low-margin tail
$\Pr[\delta_t\!<\!0]$ stays in the $5\!\sim\!22\%$ band on every active
cell at every checkpoint, while the within-bucket entropy contracts
$2$--$3\!\times$ on the productive cells (e.g.\ $A\!\cdot\!\valids$
high-entropy tail $12.0\%\!\to\!2.1\%$ between \textsc{base} and
step\,$900$).
\item $\beta\,\hat H_t$ is the \emph{entropy term}: it rewards tokens whose
local distribution is broad, independently of which token was sampled. The
moderate weight ($\beta\!=\!0.3$) keeps $c_t^{+}$ sensitive to flat-prior
regions even when $\delta_t$ is small but the token was decided in a
near-uniform alternative set.
\item $\gamma\,d_t^{+}\,\hat H_t$ is the \emph{interaction term}: it assigns
additional weight only when both diagnostics are simultaneously large. The
small coefficient ($\gamma\!=\!0.1$) keeps this cross term secondary to the
additive components while preserving sensitivity to the joint low-margin,
high-entropy regime ($d_t^{+}\!\to\!1$, $\hat H_t\!\to\!1$), which receives a
controlled multiplicative boost over either individual signal.
\end{enumerate}
Algebraically, $c_t^{+}\in[0,\,\alpha+\beta+\gamma]\!=\![0,\,1]$ with the
upper bound attained only at the joint extremum $d_t^{+}\!=\!\hat H_t\!=\!1$;
the worst-case per-token residual is therefore bounded by
$\eta\,|\hat A_i|\,\lambda_{\max}\,c_{\max}$ once $c_t^{+}$ is clamped at
$c_{\max}\!=\!0.5$ (\S\ref{app:routing}).

\subsection{Negative salience $c_t^{-}$ for negative residual segments}
\label{app:magic-neg}

For tokens in active negative-residual segments ($P_{T,S}\!=\!{-}1$) the
salience should be \emph{large} when the token is a candidate
\emph{high-probability, low-value} write -- a confidently-emitted commitment
in a peaked distribution that contributed to an undesirable trajectory.
We mirror Eq.~(\ref{eq:app-magic-pos}) with two sign-flipped substitutions.

First, the centred margin enters with the opposite sign:
\begin{equation}
d_t^{-} \;=\; \sigma\!\left(\bar\delta_t/\tau_\delta\right),\quad
\tau_\delta = 0.5,\quad d_t^{-}\!\in\!(0,1).
\label{eq:app-d-neg}
\end{equation}
Now $d_t^{-}$ saturates near $1$ for above-median margins, surfacing the
\emph{overcommitted} tokens within the segment.

Second, the entropy term is replaced by its complement, the \emph{confidence}
$\mathrm{conf}_t \!=\! 1 - \hat H_t$:
\begin{equation}
\boxed{
\begin{aligned}
c_t^{-} \;=\; I_t^{-} &\;=\; \alpha\, d_t^{-}\;+\;\beta\,\mathrm{conf}_t\;+\;\gamma\, d_t^{-}\,\mathrm{conf}_t,\\
(\alpha,\beta,\gamma) &\;=\; (0.6,\,0.3,\,0.1).
\end{aligned}
}
\label{eq:app-magic-neg}
\end{equation}
The interpretation is the dual of \S\ref{app:magic-pos}: the margin term
identifies above-segment-median commitments, the confidence term identifies
peaked distributions, and the interaction term gives a multiplicative boost
to the joint event ($d_t^{-}\!\to\!1$, $\mathrm{conf}_t\!\to\!1$) -- the
canonical ``high-probability, low-value'' write in a confidently-wrong
answer span.
$c_t^{-}\in[0,1]$, with the same residual bound as
\S\ref{app:magic-pos}.

\paragraph{Polarity symmetry.}
Equations~(\ref{eq:app-magic-pos}) and~(\ref{eq:app-magic-neg}) are related
by the substitution $(\bar\delta_t \!\mapsto\! -\bar\delta_t,\, \hat H_t
\!\mapsto\! 1\!-\!\hat H_t)$; $c_t^{+}$ and $c_t^{-}$ both live in
$[0,1]$ and share the same coefficient triple $(\alpha,\beta,\gamma)$. This
symmetry ensures that, under a fixed residual budget, positive and negative
segments with the same $|\lambda|$ induce equal-magnitude per-token residuals.
Consequently, the calibration in \S\ref{app:routing} remains invariant under
sign flips of $P_{T,S}$.

\subsection{Why margin is necessary: empirical limitations of entropy alone}
\label{app:why-margin}

\begin{figure*}[t]
\centering
\includegraphics[width=0.98\linewidth]{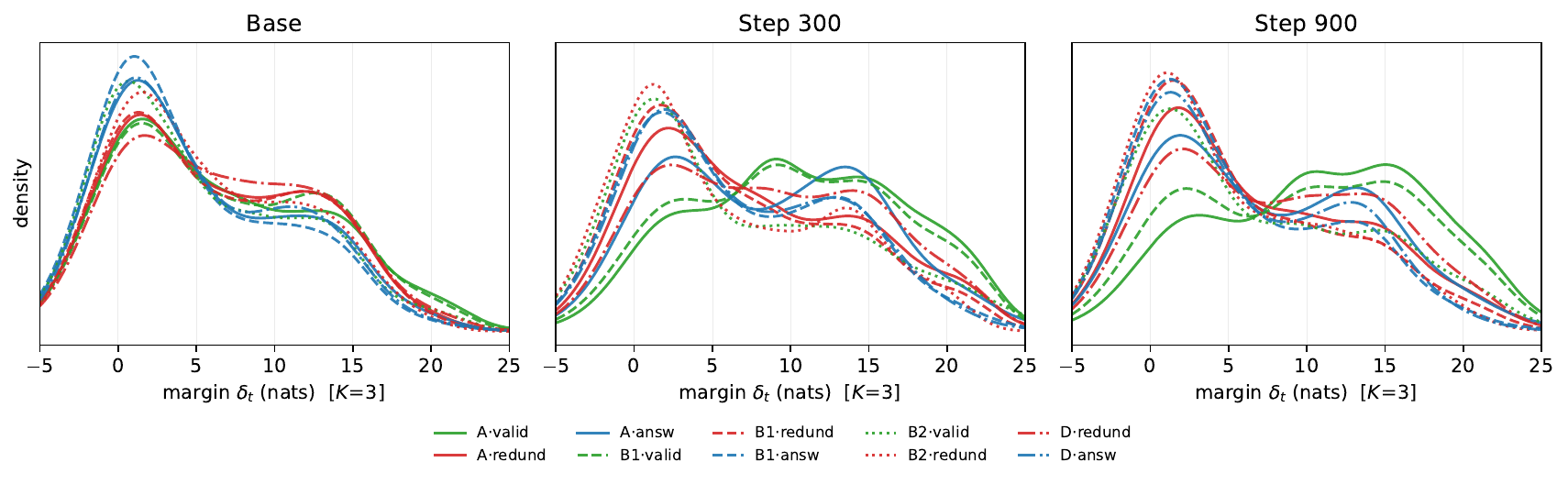}
\caption{\textbf{Density of margin@$3$ over trainable non-structural
tokens, per (T,S) bucket and checkpoint.} Line colour encodes the
segment (green: \valids, red: \redunds, blue: \answ); line style encodes
the rollout type ($A\!:\!{-}$, $B_1\!:\!\text{--}\,\text{--}$,
$B_2\!:\!\cdots$, $D\!:\!{-}\!\cdot\!{-}$).
The within-bucket distribution is heavy-tailed and overlaps substantially
with $\delta_t\!<\!0$ at every checkpoint, leaving a non-empty
\emph{low-margin} sub-population for $d_t^{+}$ to amplify on positive
segments and a heavy \emph{high-margin} sub-population for $d_t^{-}$ to
amplify on negative segments.}
\label{fig:app-margin3-distribution}
\end{figure*}

\begin{figure*}[t]
\centering
\includegraphics[width=0.98\linewidth]{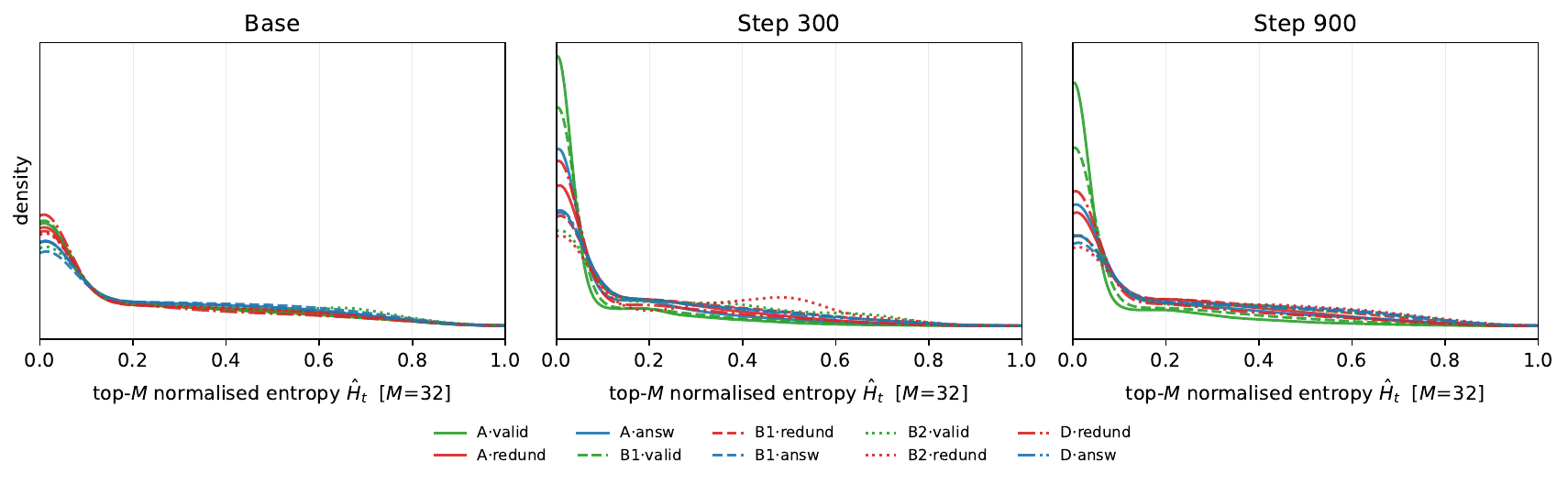}
\caption{\textbf{Density of entropy@$32$ over trainable non-structural
tokens, per (T,S) bucket and checkpoint.} Encoding matches
Fig.~\ref{fig:app-margin3-distribution}.
The mass concentrates near $\hat H_t\!\to\!0$ as training progresses
(more tokens become high-confidence picks), but a non-empty right tail
above $\hat H_t\!=\!0.5$ persists across all active buckets even at
\textsc{step\,900}, especially on the failure-mode buckets
$B_1\!\cdot\!\answ$, $B_1\!\cdot\!\redunds$, and $B_2$.}
\label{fig:app-entropy32-distribution}
\end{figure*}

The case for the additive composition in
Eqs.~(\ref{eq:app-magic-pos})--(\ref{eq:app-magic-neg}) rests on three
empirical facts about the per-segment margin and entropy distributions on
the Search-R1-trained Qwen2.5-3B run.
We summarise the bucket-level statistics in
Table~\ref{tab:app-margin-entropy-stats} and the corresponding
distributional shapes in
Figs.~\ref{fig:app-margin3-distribution}--\ref{fig:app-entropy32-distribution};
implications follow below.
Both summaries are computed over the residual-eligible tokens, namely all
trainable assistant tokens whose ID is \emph{not} in the structural-delimiter
set $\mathcal{T}_{\mathrm{struct}}$; no further per-token taxonomy is applied.

\begin{table*}[t]
\caption{\textbf{Per-(checkpoint, type, segment) margin and entropy on
trainable non-structural tokens.}
``$\bar\delta$@$3$'' is the bucket mean of the sampled-token margin
against the top-$3$ competitors; ``$\hat H$@$32$'' is the bucket mean of
the top-$32$ normalised entropy. ``low-$\delta$''~/~``high-$\hat H$''
denote tail fractions $\Pr[\delta_t\!<\!0]$ and
$\Pr[\hat H_t\!>\!0.5]$, respectively. Buckets with $P\!=\!0$
(Type-$C$, all segments) are listed for reference but contribute no
residual.}
\label{tab:app-margin-entropy-stats}
\centering
\footnotesize
\setlength{\tabcolsep}{4pt}
\begin{tabular}{@{}l|rr|rr|rr@{}}
\toprule
\multirow{2}{*}{Bucket $(T,S)$}
& \multicolumn{2}{c|}{Base} & \multicolumn{2}{c|}{Step\,300} & \multicolumn{2}{c}{Step\,900} \\
& $\bar\delta$@$3$ / $\hat H$@$32$ & low-$\delta$ / high-$\hat H$
& $\bar\delta$@$3$ / $\hat H$@$32$ & low-$\delta$ / high-$\hat H$
& $\bar\delta$@$3$ / $\hat H$@$32$ & low-$\delta$ / high-$\hat H$ \\
\midrule
$A\!\cdot\!\valids$    & 6.61 / 0.167 & 16.1\% / 12.0\% & 10.65 / 0.052 & \phantom{0}5.1\% / \phantom{0}1.4\% & 10.95 / 0.059 & \phantom{0}5.9\% / \phantom{0}2.1\% \\
$A\!\cdot\!\redunds$   & 6.22 / 0.178 & 16.7\% / 13.7\% & \phantom{0}7.55 / 0.119 & 11.9\% / \phantom{0}5.1\% & \phantom{0}6.70 / 0.152 & 15.0\% / \phantom{0}8.5\% \\
$A\!\cdot\!\answ$      & 5.35 / 0.204 & 19.7\% / 15.5\% & \phantom{0}8.73 / 0.091 & \phantom{0}9.1\% / \phantom{0}3.7\% & \phantom{0}7.44 / 0.139 & 13.7\% / \phantom{0}8.8\% \\
$B_1\!\cdot\!\valids$  & 6.72 / 0.161 & 15.5\% / 11.6\% & 10.04 / 0.068 & \phantom{0}6.6\% / \phantom{0}2.6\% & \phantom{0}9.74 / 0.089 & \phantom{0}8.9\% / \phantom{0}4.4\% \\
$B_1\!\cdot\!\redunds$ & 6.14 / 0.185 & 17.3\% / 14.7\% & \phantom{0}6.58 / 0.154 & 15.3\% / \phantom{0}8.8\% & \phantom{0}5.76 / 0.189 & 18.4\% / 12.8\% \\
$B_1\!\cdot\!\answ$    & 4.71 / 0.232 & 22.3\% / 19.0\% & \phantom{0}6.75 / 0.150 & 14.9\% / \phantom{0}8.9\% & \phantom{0}5.50 / 0.208 & 19.9\% / 16.2\% \\
$B_2\!\cdot\!\valids$  & 5.31 / 0.222 & 22.1\% / 19.8\% & \phantom{0}6.79 / 0.182 & 17.8\% / 12.3\% & \phantom{0}6.65 / 0.191 & 18.6\% / 14.7\% \\
$B_2\!\cdot\!\redunds$ & 5.70 / 0.189 & 16.9\% / 14.1\% & \phantom{0}6.00 / 0.203 & 16.0\% / 17.0\% & \phantom{0}5.30 / 0.220 & 20.7\% / 17.7\% \\
$D\!\cdot\!\redunds$   & 6.69 / 0.153 & 14.8\% / 11.2\% & \phantom{0}8.61 / 0.099 & \phantom{0}9.7\% / \phantom{0}4.6\% & \phantom{0}7.93 / 0.124 & 12.2\% / \phantom{0}7.3\% \\
$D\!\cdot\!\answ$      & 5.25 / 0.205 & 19.5\% / 15.8\% & \phantom{0}6.72 / 0.147 & 14.2\% / \phantom{0}8.3\% & \phantom{0}5.85 / 0.193 & 18.4\% / 14.8\% \\
\midrule
$C\!\cdot\!\redunds$ (ref.) & 6.06 / 0.190 & 18.0\% / 15.1\% & \phantom{0}7.72 / 0.128 & 12.5\% / \phantom{0}7.0\% & \phantom{0}6.71 / 0.169 & 16.5\% / 11.5\% \\
$C\!\cdot\!\answ$ (ref.)    & 4.52 / 0.242 & 23.2\% / 20.3\% & \phantom{0}6.07 / 0.179 & 17.3\% / 11.9\% & \phantom{0}4.87 / 0.243 & 23.4\% / 21.1\% \\
\bottomrule
\end{tabular}
\end{table*}

\paragraph{Fact 1 -- entropy contracts more aggressively than margin on
the productive cells.}
On the cells where the policy converges to confident behaviour during RL
($A\!\cdot\!\valids$, $A\!\cdot\!\answ$, $B_1\!\cdot\!\valids$,
$D\!\cdot\!\redunds$), the bucket-mean entropy contracts by $2$--$3\!\times$
between \textsc{base} and step\,$900$ (e.g.\ $A\!\cdot\!\valids$:
$0.167\!\to\!0.059$; $A\!\cdot\!\answ$: $0.204\!\to\!0.139$;
$B_1\!\cdot\!\valids$: $0.161\!\to\!0.089$), and the high-entropy tail
$\Pr[\hat H_t\!>\!0.5]$ falls correspondingly (from $\sim\!12\%$ to
$\sim\!2$--$5\%$). Margin grows in the same regime but more slowly: the
bucket mean rises by $\sim\!50$--$70\%$ on the search cells and the
low-margin tail $\Pr[\delta_t\!<\!0]$ remains in the $5$--$15\%$ band on
every cell at step\,$900$.
\textbf{Implication}: at mid-to-late training the entropy term
$\beta\,\hat H_t$ contributes shrinking density to $c_t^{\pm}$ on the
productive cells, while the margin term retains a non-trivial
discrimination range; the additive form ensures the salience score does
not degenerate as one signal contracts.

\paragraph{Fact 2 -- the failure-mode cells keep both signals alive.}
On the bottleneck cells the methodology is most concerned with --
$B_1\!\cdot\!\answ$, $B_1\!\cdot\!\redunds$, and $B_2$ -- the
distributions are markedly more heterogeneous and \emph{both} signals
remain informative at every checkpoint.
At step\,$900$, $B_1\!\cdot\!\answ$ retains a high-entropy tail of
$16.2\%$, a low-margin tail of $19.9\%$, and a bucket mean
$\bar\delta@3\!=\!5.50$ -- well below the productive-cell asymptote --
and $B_2\!\cdot\!\redunds$ shows a similar pattern ($17.7\%$ high-entropy,
$20.7\%$ low-margin).
The per-token EDA examples on these cells are coherent with the
methodology's intent: the high-$\delta$,
low-$\hat H$ region of $B_1\!\cdot\!\answ$ surfaces digit /
entity-suffix BPE pieces (`9', `8', `0', `wich', ` Alb') -- the
confidently-wrong year/entity commits the negative residual is designed
to suppress -- while the low-$\delta$, high-$\hat H$ region of
$A\!\cdot\!\valids$ surfaces query-NL words (`role', `lineage', `parent',
`quote') -- the exploration-relevant phrasing decisions the positive
residual is designed to boost.
The median-centring step in Eq.~(\ref{eq:app-centered-margin}) explicitly
normalises away the across-bucket mean shift so that $d_t^{\pm}$ remains
a \emph{within-bucket} discriminator at every checkpoint.

\paragraph{Fact 3 -- the two signals are not redundant on any single
cell.}
Within every active bucket the per-token margin and entropy are weakly
related but never collinear: e.g.\ on $B_1\!\cdot\!\answ$ at step\,$900$
the high-entropy tail ($16.2\%$) and the low-margin tail ($19.9\%$)
overlap on only $\sim\!5\%$ of tokens (Fig.~\ref{fig:app-margin3-distribution}
together with Fig.~\ref{fig:app-entropy32-distribution}: the two
distributions evolve under different dynamics).
A score using only $d_t^{\pm}$ would assign zero salience to the
high-entropy / median-margin tokens (a substantial sub-population on the
failure-mode cells), and a score using only $\hat H_t$ would assign zero
salience to the high-margin / low-entropy tokens (the dominant
sub-population on the productive cells).
\textbf{Implication}: an additive composition with comparable weights
($\alpha\!=\!0.6$, $\beta\!=\!0.3$) is the minimal aggregator that gives
non-zero salience to a token surfaced by either signal alone, and the
small interaction weight $\gamma\!=\!0.1$ multiplicatively boosts the
joint event without overwhelming the additive base.

\subsection{Why the composition identifies the right tokens}
\label{app:composition-rationale}

We now state the composition rule that follows from
Facts~1--3 and matches the polarity table of \S\ref{app:routing}.

\paragraph{P{=}{+}1 surfaces high-value low-probability tokens.}
On a positive segment $S$, $c_t^{+}$ is large iff
\emph{either} (i) $\bar\delta_t \ll 0$ (the chosen token was decided against
strong competitors -- it was not the highest-probability candidate) \emph{or}
(ii) $\hat H_t \to 1$ (the local distribution was flat, so even a
high-margin pick had a substantial alternative-token mass) -- and is
\emph{maximal} when both fire (multiplicative interaction term
$\gamma d_t^{+} \hat H_t$).
Tokens in this regime are precisely the ``high-value low-probability''
candidates the residual aims to amplify: they were valuable enough to
contribute to a productive trajectory ($P_{T,S}\!=\!{+}1$ holds at the
trajectory level), but the old policy did not prefer them strongly under its
own decoding distribution.
The EDA's top-$10$ winners by
$-\bar z(\delta_t) + \bar z(\hat H_t)$ on the active P=+1 cells are
coherent with this characterisation -- search-NL phrasing tokens
(`role', `lineage', `parent', `quote', `details') on
$A\!\cdot\!\valids$ and $B_1\!\cdot\!\valids$, and entity-choice tokens
(`John', `Manchester', ` Casino') on $A\!\cdot\!\answ$.

\paragraph{P{=}{-}1 surfaces low-value high-probability tokens.}
On a negative segment, $c_t^{-}$ is large iff $\bar\delta_t \gg 0$ (the
chosen token was strongly preferred over its top-$K$ competitors) and/or
$\mathrm{conf}_t \to 1$ (the local distribution was peaked).
Tokens in this regime are the dual: confident commitments emitted in
a trajectory that the trajectory-level signal flagged as undesirable
($P_{T,S}\!=\!-1$). Suppressing exactly these tokens is what
distinguishes \method{}'s residual from a uniform negative advantage on the
segment, which would penalise low-confidence tokens equally and risk
collateral damage to the formatting / structural surface of the response.
The EDA's top-$10$ losers by
$\bar z(\delta_t) - \bar z(\hat H_t)$ on the active P=$-1$ cells confirm
this: digit and proper-noun BPE pieces (`9', `8', `0', `wich', `zej',
`shire') on $B_1\!\cdot\!\answ$ and $B_1\!\cdot\!\redunds$ -- exactly
the confidently-wrong year and entity commitments and the over-search
filler tokens, respectively.

\paragraph{Why the additive form, not a product or a max.}
Three alternatives we considered were
(a) a pure product $c_t \!=\! d_t^{\pm}\,\hat H_t$,
(b) a pure max $c_t \!=\! \max(d_t^{\pm}, \hat H_t)$, and
(c) a margin-only $c_t\!=\!d_t^{\pm}$.
Form~(a) is sub-additive: $c_t\!\le\!\min(d_t^{\pm},\hat H_t)$, so the
score cannot exceed the smaller of the two signals on any token.
On the productive cells where the entropy mean has contracted to
$\hat H_t\!\sim\!0.06$--$0.14$ at step\,$900$ (Fact~1), the per-token
upper bound under a pure product is below $0.15$ regardless of how large
$d_t^{\pm}$ becomes, so the cell can never claim the full
$c_{\max}\!=\!0.5$ residual budget even on its highest-margin tokens.
Form~(b) is robust to per-cell shrinkage but loses the co-firing
amplification signal -- the joint $(d_t^{\pm}\!\to\!1, \hat H_t\!\to\!1)$
event identified in Fact~3 -- which the EDA's per-token examples confirm
is the most polarity-aligned region.
Form~(c) drops the entropy term entirely and is well-defined throughout
training, but leaves the high-entropy / median-margin sub-population
identified in Fact~3 (e.g.\ $\sim\!16\%$ of tokens on $B_1\!\cdot\!\answ$
at step\,$900$, and $\ge\!12\%$ of tokens on every active cell at
\textsc{base}) without any salience boost.
The additive-with-interaction form
Eqs.~(\ref{eq:app-magic-pos})--(\ref{eq:app-magic-neg}) is the simplest
composition that (i) does not degenerate when one signal vanishes,
(ii) gives the joint event a multiplicative boost, and (iii) keeps both
signals' weights interpretable as fractions of the unit budget
$\alpha+\beta+\gamma = 1$.

\subsection{Robustness of $K{=}3$ and $M{=}32$}
\label{app:topk-topm-robustness}

The salience score depends on two bandwidth hyperparameters: the number of
competitors $K$ in the margin definition (Eq.~\ref{eq:app-comp}) and the
top-$M$ truncation in the entropy definition (Eq.~\ref{eq:app-entropy}).
We argue from EDA evidence and an ablation sweep that $K\!=\!3$ and
$M\!=\!32$ are conservative defaults whose specific values are not
load-bearing.

\subsubsection{Distributional stability of margin@$K$ and entropy@$M$}
\label{app:topk-topm-stability}

For each of the $36$ EDA cells (3 checkpoints $\times$ 12 active $(T,S)$
buckets) we recompute margin@$K$ over $K\!\in\!\{2,3,5,10,32,63\}$ and
entropy@$M$ over $M\!\in\!\{2,4,8,16,32,64\}$ and track two
distribution-level summaries: the bucket mean and, for margin, the
overlap of the bottom-$1\%$ tail (the active-region for $d_t^{\pm}$)
against the $K\!=\!3$ anchor.
Table~\ref{tab:app-km-stability} reports the worst-case statistics across
the $36$ cells.

\begin{table}[H]
\caption{\textbf{Worst-case cell-level shifts of margin@$K$ and
entropy@$M$ across the $36$ EDA cells (3 checkpoints $\times$ 12 active
$(T,S)$ buckets).}
``$|\Delta\bar\delta|$'' is the maximum absolute shift of the bucket-mean
margin (in nats) relative to the $K\!=\!3$ anchor;
``min overlap@$1\%$'' is the minimum overlap of the bottom-$1\%$ margin
tail against the $K\!=\!3$ tail.
``$|\Delta\hat H|$'' is the analogous bucket-mean shift of normalised
entropy relative to the $M\!=\!32$ anchor.
Bucket-mean margin shifts smoothly and monotonically in $K$ (a known
property of $\log\mathrm{mean}\exp$ as more competitors are added), but
the active region -- the bottom-$1\%$ tail -- is preserved at $\ge\!0.75$
overlap across the entire $\sim\!30\!\times\!$ sweep, so the same
high-salience tokens absorb the residual regardless of $K$.
Bucket-mean entropy shifts are below $0.034$ for $M\!\ge\!16$ and below
$0.077$ for $M\!\ge\!8$, putting $M\!=\!32$ on a stable plateau with
$M\!=\!64$.}
\label{tab:app-km-stability}
\centering
\small
\resizebox{\columnwidth}{!}{%
\begin{tabular}{@{}c@{\hspace{1em}}c@{\hspace{1em}}c|c@{\hspace{1em}}c@{}}
\toprule
\multicolumn{3}{c|}{Margin (vs.\ $K\!=\!3$ anchor)} & \multicolumn{2}{c}{Entropy (vs.\ $M\!=\!32$ anchor)} \\
$K$ & $|\Delta\bar\delta|$ (nats) & min overlap@$1\%$ & $M$ & $|\Delta\hat H|$ \\
\midrule
$2$  & $0.343$ & $0.83$ & $2$  & $0.202$ \\
$\mathbf{3}$  & $\mathbf{0.000}$ & $\mathbf{1.00}$ & $4$  & $0.131$ \\
$5$  & $0.477$ & $0.75$ & $8$  & $0.077$ \\
$10$ & $1.157$ & $0.75$ & $16$ & $0.034$ \\
$32$ & $2.316$ & $0.75$ & $\mathbf{32}$ & $\mathbf{0.000}$ \\
$63$ & $2.993$ & $0.75$ & $64$ & $0.026$ \\
\bottomrule
\end{tabular}}
\end{table}

\begin{figure*}[t]
\centering
\includegraphics[width=0.98\linewidth]{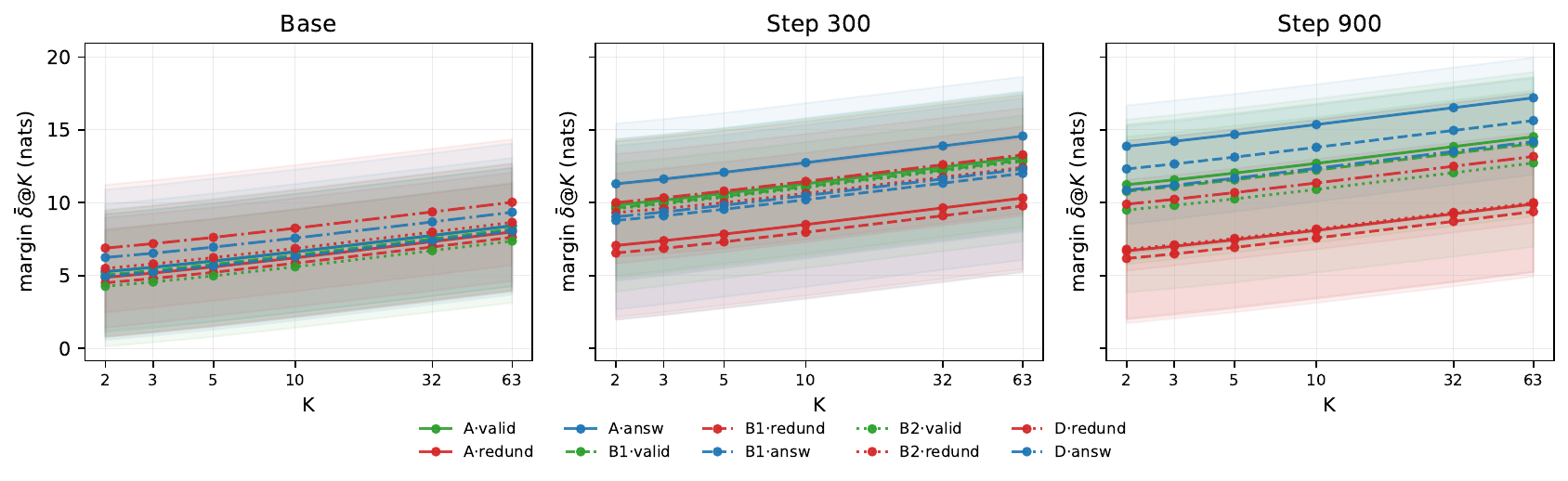}
\caption{\textbf{Bucket-conditional margin@$K$ curves with p25/p75 bands at
\textsc{base} (left), \textsc{step\,300} (middle), \textsc{step\,900}
(right).} Each panel overlays all active $P\!\neq\!0$ $(T,S)$ buckets;
line colour encodes the segment (green: \valids, red: \redunds, blue:
\answ) and line style encodes the rollout type
($A\!:\!{-}$, $B_1\!:\!\text{--}\,\text{--}$, $B_2\!:\!\cdots$,
$D\!:\!{-}\!\cdot\!{-}$).
The $K$-axis sweeps $\{2,3,5,10,32,63\}$ on a log scale.
Curves shift smoothly and monotonically with $K$ (a
$\log\mathrm{mean}\exp$ property), but the relative ordering of buckets
and the within-bucket dispersion (p25--p75 band) are preserved across the
full sweep, matching the bottom-$1\%$ overlap statistic in
Table~\ref{tab:app-km-stability}.}
\label{fig:app-margin-k-curve}
\end{figure*}

\begin{figure*}[t]
\centering
\includegraphics[width=0.98\linewidth]{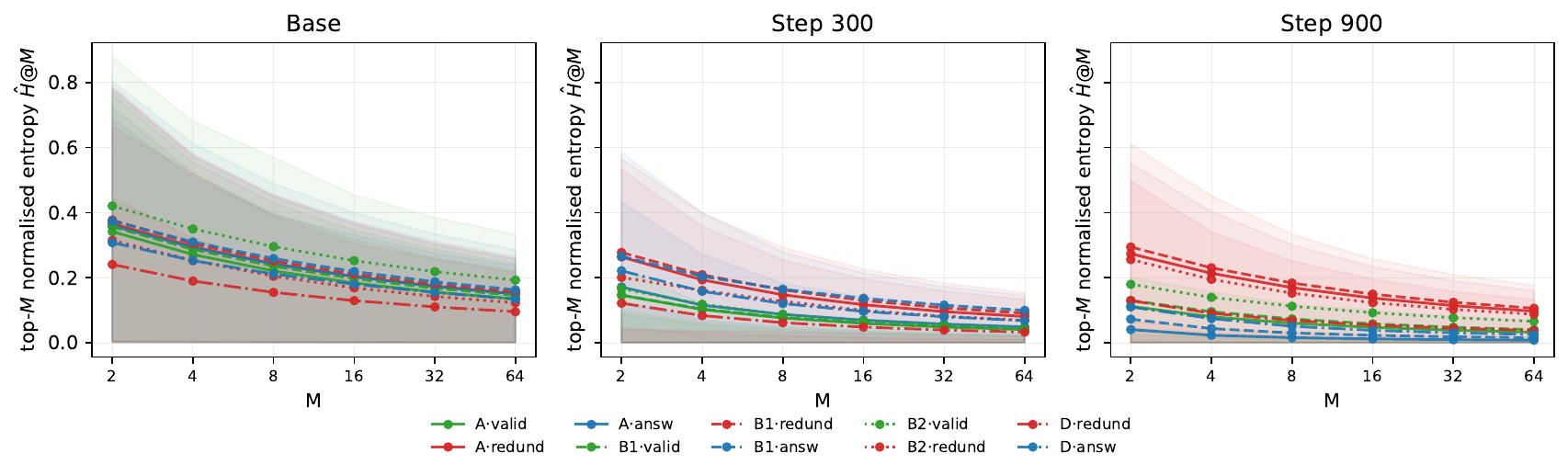}
\caption{\textbf{Bucket-conditional entropy@$M$ curves with p25/p75 bands at
\textsc{base} (left), \textsc{step\,300} (middle), \textsc{step\,900}
(right).} The $M$-axis sweeps $\{2,4,8,16,32,64\}$ on a log scale; legend
encoding matches Fig.~\ref{fig:app-margin-k-curve}.
Once $M\!\ge\!16$ the bucket means are within $0.034$ of their $M\!=\!32$
anchor and within $0.026$ of $M\!=\!64$ across all $36$ cells, justifying
$M\!=\!32$ as a stable plateau choice.}
\label{fig:app-entropy-m-curve}
\end{figure*}

\textbf{Implication}: the salience score's active-region tokens are
preserved across the swept range with $\ge\!75\%$ overlap on the
bottom-$1\%$ margin tail and a sub-$0.08$-nat shift on the entropy
mean for $M\!\ge\!8$. The choice of $K\!=\!3$ and $M\!=\!32$ is therefore
not load-bearing for which tokens absorb the residual; we adopt these
values as the operating defaults on two practical grounds.
First, $K\!=\!3$ is the smallest value at which the log-mean-exp
competitor score in Eq.~(\ref{eq:app-comp}) is supported by more than one
non-anchor competitor (and at $K\!=\!2$ the bottom-$1\%$ overlap drops
to $0.83$, just below the inner plateau), and it matches the per-step
top-$K$ logprob slice that vLLM emits at no added inference cost.
Second, $M\!=\!32$ is the smallest value at which the top-$M$
distribution captures $\ge\!99.5\%$ of probability mass on every bucket
in the EDA, ensuring $\hat H_t$ is not biased low by tail truncation;
the marginal benefit of $M\!=\!64$ over $M\!=\!32$ ($\le\!0.026$ in mean)
is below the noise floor in Table~\ref{tab:app-km-stability}.
Distributional plots of margin@$K$ and entropy@$M$ across the sweep
appear in Figs.~\ref{fig:app-margin-k-curve}--\ref{fig:app-entropy-m-curve}.

\subsubsection{End-to-end ablation on $K$ and $M$}
\label{app:topk-topm-ablation}

The distributional analysis in \S\ref{app:topk-topm-stability} establishes
that the \emph{active-region token set} and the bucket-level summaries of
margin@$K$ / entropy@$M$ are essentially invariant under variations of $K$
and $M$, but the salience \emph{score} -- specifically the
$\sigma(\bar\delta_t/\tau_\delta)$ saturation level and the
$\hat H_t / (1\!-\!\hat H_t)$ tail mass -- can shift under bandwidth changes
even when the active set does not.
We therefore complement the distributional analysis with an end-to-end
ablation:
we sweep $K\in\{2,3,5,32\}$ at fixed $M\!=\!32$ and
$M\in\{8,16,32,64\}$ at fixed $K\!=\!3$, holding all other coefficients at
the defaults of Table~\ref{tab:app-routing-final}, and report
EM, \toolhit, \hitacc, \supcorr{} on the Search-R1 evaluation suite at
checkpoint $400$.
The reported \method{} run uses $K\!=\!3,M\!=\!32$ (boldface row).

\begin{table}[H]
\caption{\textbf{Sensitivity of \method{} to the margin bandwidth $K$ on
Qwen2.5-3B} (fixed $M\!=\!32$). The bold $K\!=\!3$ row is the reported
configuration. EM deviates by at most $0.34$ points from the default
$K\!=\!3$ setting; the full range across tested settings is $0.46$ points.
All four metrics remain within $1.0$ point of the $K\!=\!3$ anchor.}
\label{tab:app-topk-ablation}
\centering
\small
\begin{tabular}{@{}lcccc@{}}
\toprule
$K$ & EM & \toolhit & \hitacc & \supcorr \\
\midrule
$2$  & 42.72 & 64.26 & 66.34 & 42.63 \\
$\mathbf{3}$  & \textbf{43.06} & \textbf{64.68} & \textbf{66.17} & \textbf{42.80} \\
$5$  & 42.78 & 64.06 & 66.13 & 42.36 \\
$32$ & 43.18 & 64.96 & 65.29 & 42.41 \\
\bottomrule
\end{tabular}
\end{table}

\begin{table}[H]
\caption{\textbf{Sensitivity of \method{} to the entropy bandwidth $M$ on
Qwen2.5-3B} (fixed $K\!=\!3$). The bold $M\!=\!32$ row is the reported
configuration. EM deviates by at most $0.75$ points from the default
$M\!=\!32$ setting (at $M\!=\!64$); the full range across tested settings
is $0.79$ points. All four metrics remain within $1.0$ point of the
$M\!=\!32$ anchor.}
\label{tab:app-topm-ablation}
\centering
\small
\begin{tabular}{@{}lcccc@{}}
\toprule
$M$ & EM & \toolhit & \hitacc & \supcorr \\
\midrule
$8$  & 42.96 & 64.08 & 65.79 & 42.16 \\
$16$ & 43.10 & 64.50 & 66.03 & 42.58 \\
$\mathbf{32}$ & \textbf{43.06} & \textbf{64.68} & \textbf{66.17} & \textbf{42.80} \\
$64$ & 42.31 & 64.10 & 65.93 & 42.26 \\
\bottomrule
\end{tabular}
\end{table}

\paragraph{Outcome.}
The end-to-end ablation is consistent with the distributional robustness
analysis in \S\ref{app:topk-topm-stability}.
Across the $K$ sweep, EM ranges from $42.72$ to $43.18$ (a range of
$0.46$ points across settings and a maximum deviation of $0.34$ from
$K\!=\!3$), while the ranges of \toolhit{}, \hitacc{}, and \supcorr{}
across settings are $0.90$, $1.05$, and $0.44$ points, respectively; all
settings remain close to the $K\!=\!3$ anchor and show no monotone
dependence on $K$.
For $M$, the inner plateau $M\!\in\!\{16,32\}$ is nearly flat
($\le\!0.04$ EM and $\le\!0.22$ \supcorr{} gap), whereas $M\!=\!64$
incurs a small but consistent drop relative to $M\!=\!32$ ($0.75$ EM
and $0.54$ \supcorr{}; the EM range across all tested $M$ is $0.79$),
suggesting that very large entropy supports can overweight low-mass
rare-token positions.
These results support $(K,M)\!=\!(3,32)$ as the operating default and
indicate that further tuning is better allocated to $\eta$,
$\lambda_{T,S}$, or $c_{\max}$ (\S\ref{app:sensitivity}).



\section{Oracle-query retrievability}
\label{app:retrievability}

For \toolhit{}$=0$ rollouts (types~$C$ and~$D$), we separate \emph{actor failure} (ineffective queries) from \emph{corpus absence} (the answer is not covered by the retrieval corpus).
We instantiate templated oracle queries from the gold answer aliases and submit them to the same retrieval pipeline.
If the oracle queries also fail to retrieve the answer, the question is treated as a corpus miss.
To avoid false negatives, we conservatively set $\Retr(x)=1$ whenever any rollout in the group has $H_i=1$, regardless of the oracle-query result.
This check affects only type-$D$ rollouts: retrievable type-$D$ rollouts receive the residual penalty for unsupported correctness (\S\ref{sec:routing}), whereas non-retrievable ones are masked, so that corpus coverage gaps are not conflated with retrieval-policy errors.
Masking acts only on the policy loss: masked rollouts still contribute to the group mean and standard deviation, so the advantages of the other rollouts are unaffected.
In our training logs, type-$D$ rollouts account for 2--3\% of all rollouts, groups that are entirely type $D$ with zero reward variance for 0.5\% of groups, and groups consisting only of non-retrievable type-$D$ rollouts for under 0.1\%.

\section{Reliability of the alias-based grounding signal}
\label{app:alias-audit}

$H_i$ is computed on the tool responses actually received by the policy: the retriever returns the top-3 passages and the response is truncated to 1,024 tokens before matching, so an alias in a truncated passage is missed.
Matching is a case-insensitive substring test, and aliases are combined disjunctively as in the EM reward.
Table~\ref{tab:alias-audit} lists potentially problematic cases over all 169,615 training questions (79,168 NQ and 90,447 HotpotQA).

\begin{table}[H]
\caption{\textbf{Audit of gold aliases in the training data.}}
\label{tab:alias-audit}
\centering
\small
\setlength{\tabcolsep}{3pt}
\begin{tabular}{@{}lrr@{}}
\toprule
Case & Count & Share \\
\midrule
Yes/no gold answer (5,481 in HotpotQA) & 5,483 & 3.23\% \\
More than one alias (all NQ) & 7,750 & 4.57\% \\
Alias of $\le4$ characters & 26,595 & 15.68\% \\
\quad of which $\le2$ characters & -- & 3.30\% \\
Single-token alias & 56,004 & 33.02\% \\
Numeric-only alias & 16,198 & 9.55\% \\
Formatting artefacts (e.g., \texttt{[4]}, NBSP) & 1,640 & 0.97\% \\
\bottomrule
\end{tabular}
\end{table}

\paragraph{Recall on annotated evidence.}
We apply the same matcher to the annotated supporting documents of the 90,447 HotpotQA training questions (53.3\% of the training data).
The alias is absent in 3,269 questions (3.61\%); excluding yes/no answers, it is absent in 34 of 84,966 (0.040\%), all formatting artefacts that Unicode or punctuation normalisation would recover.
This shows high recall on genuine supporting evidence for ordinary answers; it does not measure precision on arbitrary retrieved passages, where an answer string may appear without supporting the answer.
Yes/no answers are inferred verdicts and would require an evidence-entailment criterion instead.

\paragraph{Removing yes/no questions.}
We retrained Qwen2.5-3B-Instruct without yes/no questions under otherwise identical settings.
After 100 training steps, in-domain EM on NQ and HotpotQA improves by about 0.7 points, while performance on the five out-of-domain benchmarks is approximately unchanged.
As a single-seed, 100-step comparison, this result is only indicative.

\section{Ablation and learned-routing details}
\label{app:extra-ablation}

\paragraph{Ablation details.}
For \emph{w/o salience}, each token-dependent $c_{i,t}$ is replaced by a cell-specific constant $\bar c_{T,S}$ measured on a frozen \method rollout sample, which preserves each cell's average residual budget.
For \emph{w/o segment split}, \valids{} and \redunds{} are merged into one search segment per type; its polarity and magnitude are fixed before training from the token-weighted signed and absolute residual means of the two segments.
Answer routing, trajectory diagnosis, data, and optimisation settings are unchanged; results are in Table~\ref{tab:ablation}.

\paragraph{Learned routing magnitudes.}
The constrained router keeps the fixed polarity $P_{T,S}$ and learns each active cell's magnitude $\lambda_{T,S}\in[0,1]$, with the same trajectory diagnosis, salience, and bound $|\Delta A_{i,t}|\le0.1|\hat A_i|$.
Router logits are initialised from the fixed table and frozen for the first 20 actor updates; they are then updated every 5 actor steps with AdamW (learning rate $5{\times}10^{-5}$, cosine schedule, weight decay $10^{-4}$, gradient-norm cap 1.0) to maximise EM $+\,0.25\,\supcorr$ on a stratified 1\% NQ--HotpotQA calibration split.
After each update, the magnitudes are projected onto a type-wise capped simplex that preserves the original residual budget.
Results over three seeds are in Table~\ref{tab:learned}.

\section{Training and evaluation hyperparameters}
\label{app:hparams}

We use the same training and evaluation splits as Search-R1~\citep{jin2025searchr1} and otherwise follow its training recipe, modifying only the token-level loss.
Tool invocation uses the Hermes-native tool-calling interface.
For Qwen2.5-3B-Instruct (GRPO), we use a batch size of $512$ prompts, $5$ rollouts per prompt, learning rate $1\!\times\!10^{-6}$, clip ratio $\epsilon=0.2$, max assistant turns $5$, and max generation length $768$ per turn.
Search-R1 reproduction runs and other reproduced baselines use KL coefficient $\beta=0.001$, while \method runs use $\beta=0.005$.
For Qwen3-30B-A3B-Instruct-2507 (GSPO), we use the standard sequence-level ratio and sequence-level clipping with the same $\epsilon$.
Training rollouts are sampled with temperature $1.0$, following Search-R1, whereas QA evaluation uses temperature $0.0$.
Training and evaluation share the same retrieval pipeline, including the retriever, corpus, retrieval depth, and tool-response truncation (an E5~\citep{wang2022e5} retriever over the 2018 Wikipedia dump, top-$3$ passages per query, at most $1024$ tool-response tokens), as well as the same agent loop (at most $2$ parallel tool calls per turn, max generation length $768$, and at most $5$ turns).
Experiments use $128$ H20 GPUs in total, organised into two experiment pools, with an $8$-H20 retrieval-service pool to support high-concurrency retrieval.

\section{Search-R1 KL control}
\label{app:searchr1-kl}

Because our main \method runs use a larger KL coefficient than the reproduced Search-R1 baseline, we additionally train Search-R1 with $\beta=5\!\times\!10^{-3}$ while keeping all other settings unchanged.

\begin{table}[H]
\caption{\textbf{Search-R1 KL control on Qwen2.5-3B-Instruct.} Aggregate percentages over the seven QA benchmarks.}
\label{tab:searchr1-kl-control}
\centering
\small
\setlength{\tabcolsep}{1.6pt}
\begin{tabular}{@{}lccccc@{}}
\toprule
KL coef. & EM & \toolhit & \hitacc & \supcorr & \supcorr{}/Acc \\
\midrule
$1\!\times\!10^{-3}$ & 37.36 & 45.79 & 65.98 & 30.21 & 76.9\% \\
$5\!\times\!10^{-3}$ & 36.61 & 45.45 & 63.31 & 28.77 & 74.4\% \\
\bottomrule
\end{tabular}
\end{table}

The higher-KL control does not improve over the default Search-R1 setting, so increasing the KL coefficient alone is not responsible for the gains of \method.

\section{Per-benchmark detail}
\label{app:detail}

\begin{table*}[t]
\caption{\textbf{Per-benchmark grounding results on Qwen2.5-3B-Instruct.} Values are percentages. \emph{Initial checkpoint}: the instruction-tuned model before task RL, evaluated with the same search tool and protocol; its EM values form the initial-checkpoint row of Table~\ref{tab:main-em}. \hitacc{}, \supcorr{}, and \supcorr{}/Acc use the lenient evaluation correctness $\tilde C$ (\S\ref{sec:exp-setup}), so \supcorr{} can exceed EM. From Search-R1 to \method, \hitacc{} falls on NQ, TriviaQA, PopQA, and HotpotQA and rises on the other three benchmarks; since the additional hits may involve questions of different difficulty, these changes alone do not indicate better or worse evidence use.}
\label{tab:full-3b}
\centering
\small
\setlength{\tabcolsep}{4pt}
\begin{tabular}{ll@{\ \ }ccccc}
\toprule
Benchmark & Method & EM & \toolhit & \hitacc & \supcorr & \supcorr{}/Acc \\
\midrule
\multirow{3}{*}{NQ}
  & Initial checkpoint & 16.15 & 59.92 & 60.80 & 36.43 & 89.3\% \\
  & Search-R1          & 47.17 & 70.33 & \textbf{67.15} & 47.23 & 97.4\% \\
  & +\method           & \textbf{48.06} & \textbf{77.76} & 63.09 & \textbf{49.06} & \textbf{99.4\%} \\
\midrule
\multirow{3}{*}{TriviaQA}
  & Initial checkpoint & 25.81 & 60.20 & 75.26 & 45.31 & 86.1\% \\
  & Search-R1          & 62.72 & 73.58 & \textbf{87.63} & 64.48 & 95.5\% \\
  & +\method           & \textbf{63.68} & \textbf{80.24} & 84.72 & \textbf{67.98} & \textbf{99.2\%} \\
\midrule
\multirow{3}{*}{PopQA}
  & Initial checkpoint & 16.54 & 43.92 & 64.25 & 28.22 & 88.0\% \\
  & Search-R1          & 40.25 & 54.90 & \textbf{74.15} & 40.71 & 97.5\% \\
  & +\method           & \textbf{42.94} & \textbf{64.39} & 69.66 & \textbf{44.85} & \textbf{99.8\%} \\
\midrule
\multirow{3}{*}{HotpotQA}
  & Initial checkpoint & 12.37 & 42.31 & 51.74 & 21.89 & 76.8\% \\
  & Search-R1          & 36.37 & 46.85 & \textbf{64.49} & 30.21 & 79.3\% \\
  & +\method           & \textbf{44.65} & \textbf{69.60} & 62.44 & \textbf{43.46} & \textbf{93.6\%} \\
\midrule
\multirow{3}{*}{2Wiki}
  & Initial checkpoint & 10.66 & 33.64 & 51.99 & 17.49 & 46.7\% \\
  & Search-R1          & 34.94 & 33.75 & 60.82 & 20.53 & 55.1\% \\
  & +\method           & \textbf{44.12} & \textbf{60.41} & \textbf{64.25} & \textbf{38.81} & \textbf{79.2\%} \\
\midrule
\multirow{3}{*}{MuSiQue}
  & Initial checkpoint & 4.18 & 25.78 & 35.15 & 9.06 & 88.3\% \\
  & Search-R1          & 12.08 & 21.93 & 45.09 & 9.89 & 71.8\% \\
  & +\method           & \textbf{18.74} & \textbf{47.54} & \textbf{46.30} & \textbf{22.01} & \textbf{98.4\%} \\
\midrule
\multirow{3}{*}{Bamboogle}
  & Initial checkpoint & 12.00 & 35.20 & 47.73 & 16.80 & 84.0\% \\
  & Search-R1          & 28.00 & 19.20 & 62.50 & 12.00 & 42.9\% \\
  & +\method           & \textbf{39.20} & \textbf{52.80} & \textbf{72.73} & \textbf{38.40} & \textbf{96.0\%} \\
\bottomrule
\end{tabular}
\end{table*}

\begin{table*}[t]
\caption{\textbf{Per-benchmark grounding results on Qwen3-30B-A3B-Instruct-2507.} Rows and metrics as in Table~\ref{tab:full-3b}.}
\label{tab:full-30b}
\centering
\small
\setlength{\tabcolsep}{4pt}
\begin{tabular}{ll@{\ \ }ccccc}
\toprule
Benchmark & Method & EM & \toolhit & \hitacc & \supcorr & \supcorr{}/Acc \\
\midrule
\multirow{3}{*}{NQ}
  & Initial checkpoint & 30.44 & 66.54 & 65.40 & 43.52 & 96.5\% \\
  & Search-R1          & 49.61 & 79.75 & 63.42 & 50.58 & 99.2\% \\
  & +\method           & \textbf{54.71} & \textbf{83.32} & \textbf{66.86} & \textbf{55.71} & \textbf{99.5\%} \\
\midrule
\multirow{3}{*}{TriviaQA}
  & Initial checkpoint & 55.57 & 71.19 & 88.09 & 62.71 & 96.0\% \\
  & Search-R1          & 69.04 & 80.58 & 88.66 & 71.44 & 98.3\% \\
  & +\method           & \textbf{71.83} & \textbf{83.87} & \textbf{90.35} & \textbf{75.78} & \textbf{99.2\%} \\
\midrule
\multirow{3}{*}{PopQA}
  & Initial checkpoint & 34.55 & 57.93 & \textbf{74.30} & 43.04 & 98.4\% \\
  & Search-R1          & 46.74 & 65.70 & 73.20 & 48.09 & 99.4\% \\
  & +\method           & \textbf{50.12} & \textbf{72.19} & 72.56 & \textbf{52.38} & \textbf{99.8\%} \\
\midrule
\multirow{3}{*}{HotpotQA}
  & Initial checkpoint & 36.25 & 62.21 & 63.56 & 39.54 & 90.7\% \\
  & Search-R1          & 45.70 & 68.71 & 62.38 & 42.86 & 92.4\% \\
  & +\method           & \textbf{52.64} & \textbf{76.41} & \textbf{67.94} & \textbf{51.91} & \textbf{94.1\%} \\
\midrule
\multirow{3}{*}{2Wiki}
  & Initial checkpoint & 36.55 & 52.46 & 66.73 & 35.01 & 74.5\% \\
  & Search-R1          & 46.14 & 58.46 & 65.57 & 38.33 & 78.4\% \\
  & +\method           & \textbf{54.18} & \textbf{66.05} & \textbf{73.25} & \textbf{48.38} & \textbf{80.1\%} \\
\midrule
\multirow{3}{*}{MuSiQue}
  & Initial checkpoint & 14.48 & 40.46 & 46.83 & 18.95 & 94.8\% \\
  & Search-R1          & 19.57 & 44.81 & 46.72 & 20.94 & 97.7\% \\
  & +\method           & \textbf{27.84} & \textbf{57.26} & \textbf{53.47} & \textbf{30.62} & \textbf{98.8\%} \\
\midrule
\multirow{3}{*}{Bamboogle}
  & Initial checkpoint & 42.40 & 49.60 & 80.65 & 40.00 & 86.2\% \\
  & Search-R1          & 47.20 & 57.60 & 76.39 & 44.00 & 93.2\% \\
  & +\method           & \textbf{56.80} & \textbf{64.00} & \textbf{83.75} & \textbf{53.60} & \textbf{94.4\%} \\
\bottomrule
\end{tabular}
\end{table*}

\section{BFCL v3 function-calling evaluation}
\label{app:bfcl}

To check whether gains on search-agent training preserve general tool-use ability, we also evaluate the Qwen3-30B-A3B-Instruct-2507 variants on BFCL v3.
Following the BFCL v3 protocol, we run the official \texttt{non\_live}, \texttt{live}, and \texttt{multi\_turn} collections with temperature $0.7$ and report the overall score using the official BFCL v3 aggregation.

\begin{table*}[t]
\caption{\textbf{BFCL v3 function-calling results on Qwen3-30B-A3B-Instruct-2507.} Values are percentages.}
\label{tab:bfcl-v3}
\centering
\small
\setlength{\tabcolsep}{5pt}
\begin{tabular}{lccccc}
\toprule
Method & Non-Live & Live & Multi-Turn & Irrelevance & v3 Overall \\
\midrule
Base      & 87.02 & \textbf{77.21} & 37.62 & \textbf{85.42} & 67.29 \\
Search-R1 & \textbf{87.71} & 77.03 & 41.12 & 85.00 & 68.62 \\
+\method  & 86.85 & 76.45 & \textbf{44.12} & 83.33 & \textbf{69.14} \\
\bottomrule
\end{tabular}
\end{table*}

The main improvement comes from multi-turn function calling, where \method increases accuracy from $41.12\%$ to $44.12\%$ over Search-R1 and by $6.50$ points over the base model.

\section{Design-space comparison}
\label{app:positioning}

\begin{table*}[t]
\caption{\textbf{Design-space comparison} with representative credit-assignment methods for search-agent RL.}
\label{tab:positioning}
\centering\small
\begin{tabular}{lccccc}
\toprule
Method & Credit granularity & Decouple S/A & W/o extra model & W/o extra rollout & One-pass \\
\midrule
Search-R1         & trajectory & \texttimes & \checkmark & \checkmark & \checkmark \\
DeSA              & trajectory & \checkmark\,(staged) & \checkmark & \checkmark & \texttimes \\
IGPO              & turn       & \texttimes & \checkmark & \checkmark & \checkmark \\
GiGPO             & step/action & \texttimes & \checkmark & \checkmark & \checkmark \\
HGPO              & step/action hier. & \texttimes & \checkmark & \checkmark & \checkmark \\
MT-PPO/GRPO  & token (GAE) & \texttimes & \texttimes\,(critic) & \texttimes\,$O(G^{K-1})$ & \texttimes \\
RTT               & token      & \texttimes & \texttimes\,(disc.) & \checkmark & \checkmark \\
\textbf{\method}  & \textbf{token} & \textbf{\checkmark}\,(routing) & \textbf{\checkmark} & \textbf{\checkmark} & \textbf{\checkmark} \\
\bottomrule
\end{tabular}
\end{table*}

Table~\ref{tab:positioning} summarises the design choices discussed in Section~\ref{sec:related}: \method is the only method in this comparison that assigns token-level credit, separates search from answer credit, and trains end-to-end in a single pass without an auxiliary model or extra rollouts.

\clearpage
%

\section{Representative Rollouts by Type}
\label{app:rollout-examples}

This appendix concretises the rollout-type taxonomy of \S\ref{sec:prelim}
and the routing matrix of Table~\ref{tab:app-routing-final} with one
representative trajectory per type.
For every example we report (i)~the question and reference answer,
(ii)~the agent's per-turn output (tool calls and final answer), and
(iii)~a compressed view of the corresponding tool response.
Tool responses are truncated to the most relevant portion of the top-$3$
documents returned by the E5 retriever; 

All rollouts share the system / user prompt template fixed by the
Search-R1 evaluation pipeline:

\begin{tcolorbox}[colback=gray!5!white, colframe=gray!50!black,
  title=System / user prompt (shared across all examples), boxrule=0.3mm,
  arc=2mm, breakable, fonttitle=\bfseries\small, fontupper=\small]
\textbf{System:} You are a helpful and harmless assistant.

\textbf{User:} Answer the given question. You must follow this interaction
pattern:
\begin{enumerate}[leftmargin=1.3em, topsep=1pt, itemsep=0pt]
\item Reason about the question inside \texttt{<think>} and
\texttt{</think>} tags.
\item If you need more information, call the search tool to retrieve
relevant knowledge.
\item After receiving search results, reason again inside
\texttt{<think>} and \texttt{</think>} tags.
\item Repeat steps 2--3 as needed. You can search multiple times.
\item When you have enough information, provide your final answer inside
\texttt{<answer>} and \texttt{</answer>} tags, without detailed
illustrations. For example, \texttt{<answer> Beijing </answer>}.
\end{enumerate}
Question: \emph{$\langle$question$\rangle$}
\end{tcolorbox}

\noindent
The colour convention in the trajectory boxes that follow:
\textcolor{deepblue}{\texttt{<tool\_call>}} and the JSON arguments are
the agent-emitted tool call;
\textcolor{deeppurple}{\texttt{<answer>}} delimits the final answer span;
agent boxes are rendered in blue and tool/observation boxes in orange.
For brevity we omit the \texttt{<|im\_start|>}/\texttt{<|im\_end|>}
turn-boundary tokens; they are present in the trained sequence.

\subsection{Type $A$: hit-correct ($H{=}1, C{=}1$)}
\label{app:example-A}

\begin{tcolorbox}[colback=gray!5!white, colframe=gray!60!black,
  title=Diagnosis, boxrule=0.3mm, arc=2mm, breakable,
  fonttitle=\bfseries\small, fontupper=\small]
\textbf{Question} (popqa, id\,57): \emph{Who is the author of A Theft?}\quad
\textbf{Reference}: \texttt{Saul Bellow}.\quad
\textbf{Predicted}: \texttt{Saul Bellow} (EM\,$=\!1$).\quad
\textbf{Pattern}: a single search returns the title-matched Wikipedia
passage in Doc\,1 and the agent extracts the author from the first sentence.
This is the modal Type-$A$ behaviour: \valids{} is the first turn and
\answ{} is the second; under the routing matrix both segments receive
$P\!=\!{+}1$ and the residual reinforces the productive search and
correct extraction.
\end{tcolorbox}

\begin{tcolorbox}[colback=blue!2!white, colframe=blue!75!black,
  title=Agent (turn 0 -- valid\_search), boxrule=0.3mm, arc=2mm,
  breakable, fonttitle=\bfseries\small, fontupper=\ttfamily\small]
\textcolor{deepblue}{<tool\_call>}\\
\{"name": "search", "arguments":\\
\quad \{"query\_list": ["Who is the author of A Theft?",\\
\qquad\qquad\qquad\quad\ "book A Theft author"]\}\}\\
\textcolor{deepblue}{</tool\_call>}
\end{tcolorbox}

\begin{tcolorbox}[colback=orange!2!white, colframe=orange!75!black,
  title={Tool response (turn 0 -- top-1 doc, truncated)},
  boxrule=0.3mm, arc=2mm, breakable, fonttitle=\bfseries\small,
  fontupper=\small]
\textbf{Doc 1 (Title: ``A Theft'')}\\
A Theft is a 1989 novel by the American author \textbf{Saul Bellow}.
Bellow originally wanted to publish the book as a story or serial in a
magazine such as ``The New Yorker'', but his agent had trouble selling it
to any magazine. Bellow, instead, chose to publish it as a book\ldots
\quad\\
\textit{[Doc 2--3 omitted: unrelated documents on ``The Sacred Art of
Stealing'' and a 1956 caper film.]}
\end{tcolorbox}

\begin{tcolorbox}[colback=blue!2!white, colframe=blue!75!black,
  title=Agent (turn 1 -- answer), boxrule=0.3mm, arc=2mm,
  breakable, fonttitle=\bfseries\small, fontupper=\small]
Based on the search result, the novel ``A Theft'' was written by the
American author Saul Bellow.\\[2pt]
\textcolor{deeppurple}{<answer>}\,Saul Bellow\,\textcolor{deeppurple}{</answer>}
\end{tcolorbox}

\subsection{Type $B_1$: hit-wrong ($H{=}1, C{=}0$, valid wrong answer)}
\label{app:example-B1}

\begin{tcolorbox}[colback=gray!5!white, colframe=gray!60!black,
  title=Diagnosis, boxrule=0.3mm, arc=2mm, breakable,
  fonttitle=\bfseries\small, fontupper=\small]
\textbf{Question} (nq, id\,7): \emph{what color is the cue ball in pool?}\quad
\textbf{Reference}: \texttt{white}.\quad
\textbf{Predicted}: \texttt{red} (EM\,$=\!0$).\quad
\textbf{Pattern}: retrieval succeeds (the top-$3$ documents do contain the
white-cue-ball fact, embedded in passages about televised
``measle'' cue balls and kaisa-billiards object-balls), but the agent
mis-aggregates the evidence and commits to ``red'' -- a token taken from
the snippet ``the \textbf{is usually red}'' (referring to a kaisa
\emph{object} ball, not a cue ball).
This is the canonical Type-$B_1$ failure mode: \emph{evidence acquisition
succeeded, evidence utilisation failed}, and the routing matrix assigns
$P_{B_1,\valids}\!=\!{+}1$ (preserve the useful search) and
$P_{B_1,\answ}\!=\!{-}1$ (suppress the confident-but-wrong answer span).
\end{tcolorbox}

\begin{tcolorbox}[colback=blue!2!white, colframe=blue!75!black,
  title=Agent (turn 0 -- valid\_search), boxrule=0.3mm, arc=2mm,
  breakable, fonttitle=\bfseries\small, fontupper=\ttfamily\small]
\textcolor{deepblue}{<tool\_call>}\\
\{"name": "search", "arguments":\\
\quad \{"query\_list": ["what color is the cue ball in pool",\\
\qquad\qquad\qquad\quad\ "color of cue ball in pool"]\}\}\\
\textcolor{deepblue}{</tool\_call>}
\end{tcolorbox}

\begin{tcolorbox}[colback=orange!2!white, colframe=orange!75!black,
  title=Tool response (turn 0 -- relevant snippets), boxrule=0.3mm,
  arc=2mm, breakable, fonttitle=\bfseries\small, fontupper=\small]
\textbf{Doc 1 (Title: ``Billiard ball'')}: \ldots Television was also the
genesis of the \textbf{cue ball with six red spots} on its surface
(known as a ``measle'' cue ball)\ldots\\
\textbf{Doc 2 (Title: ``Cue sports'')}: \ldots In kaisa, five balls are
used: the yellow (called the ``kaisa'' in Finnish), two red object balls,
and the \textbf{two white cue balls} (usually differentiated by one cue
ball having a dot or other marking on it)\ldots American-style \ldots\\
\textbf{Doc 3 (Title: ``Cue sports'')}: \ldots The cloth of the billiard
table has traditionally been green\ldots
\end{tcolorbox}

\begin{tcolorbox}[colback=blue!2!white, colframe=blue!75!black,
  title={Agent (turn 1 -- answer, \emph{valid wrong})},
  boxrule=0.3mm, arc=2mm, breakable, fonttitle=\bfseries\small,
  fontupper=\small]
Based on the search results, the cue ball in American-style pocket
billiards is usually red. This information comes from resources discussing
various types of cue balls used in different cue sports like billiards
and pool.\\[2pt]
\textcolor{deeppurple}{<answer>}\,red\,\textcolor{deeppurple}{</answer>}
\end{tcolorbox}

\subsection{Type $B_2$: hit-timeout ($H{=}1, C{=}0$, max-turn termination)}
\label{app:example-B2}

\begin{tcolorbox}[colback=gray!5!white, colframe=gray!60!black,
  title=Diagnosis, boxrule=0.3mm, arc=2mm, breakable,
  fonttitle=\bfseries\small, fontupper=\small]
\textbf{Question} (musique, id\,60): \emph{What is the name of the castle
found in the birthplace of the performer of The Speckless Sky?}\quad
\textbf{Reference}: \texttt{Casa Loma}.\quad
\textbf{Predicted}: (truncated partial -- the rollout was terminated by
the $5$-turn cap before an \texttt{<answer>\ldots</answer>} span was
emitted, EM\,$=\!0$).\quad
\textbf{Pattern}: a 2-hop question (\emph{performer of $X$} $\to$
\emph{birthplace of performer} $\to$ \emph{castle in birthplace}). The
agent solves the first hop in turn $1$, the second hop in turn $4$, and is
on the right track in turn $5$ (Doc\,1 of the final response surfaces
``Casa Loma'') but the turn budget is exhausted before the agent commits.
The bottleneck is \emph{post-hit redundant search}: the routing matrix
assigns $P_{B_2,\redunds}\!=\!{-}1$ to suppress the over-search that
consumed the budget; the per-cell $\lambda_{B_2,\redunds}\!=\!0.1$ is
deliberately small after calibration, both because $B_2$ token mass is
small (Table~\ref{tab:app-token-counts}) and because larger values
empirically interact with the $B_1$ answer-suppression gradient.
\end{tcolorbox}

\begin{tcolorbox}[colback=blue!2!white, colframe=blue!75!black,
  title={Agent (turn 0 -- redundant\_search, no hit)}, boxrule=0.3mm,
  arc=2mm, breakable, fonttitle=\bfseries\small, fontupper=\ttfamily\small]
\textcolor{deepblue}{<tool\_call>}\,\{"query\_list": ["What is the name of
the castle found in the birthplace of the performer of The Speckless Sky",
"The Speckless Sky performer"]\}\,\textcolor{deepblue}{</tool\_call>}
\end{tcolorbox}

\begin{tcolorbox}[colback=orange!2!white, colframe=orange!75!black,
  title=Tool response (turn 0 -- non-relevant), boxrule=0.3mm, arc=2mm,
  breakable, fonttitle=\bfseries\small, fontupper=\small]
\textbf{Doc 1 (Title: ``Knapp's Castle'')}: a privately owned ruined
mansion near Santa Barbara\ldots\quad
\textit{[Doc 2--3 omitted: unrelated castle articles.]}
\end{tcolorbox}

\begin{tcolorbox}[colback=blue!2!white, colframe=blue!75!black,
  title={Agent (turn 1 -- valid\_search, hit on hop $1$)}, boxrule=0.3mm,
  arc=2mm, breakable, fonttitle=\bfseries\small, fontupper=\small]
Based on the search results, the question asks about the birthplace of the
performer of ``The Speckless Sky''. Let me find the identity of the
performer of this song.\\[2pt]
\texttt{\textcolor{deepblue}{<tool\_call>}\,\{"query\_list": ["performer
of The Speckless Sky", "song The Speckless Sky"]\}\,\textcolor{deepblue}{</tool\_call>}}
\end{tcolorbox}

\begin{tcolorbox}[colback=orange!2!white, colframe=orange!75!black,
  title=Tool response (turn 1 -- hit), boxrule=0.3mm, arc=2mm,
  breakable, fonttitle=\bfseries\small, fontupper=\small]
\textbf{Doc 1 (Title: ``The Speckless Sky'')}: The Speckless Sky is an
album by \textbf{Jane Siberry}. It was Siberry's highest-charting album
on the Canadian charts\ldots
\end{tcolorbox}

\begin{tcolorbox}[colback=blue!2!white, colframe=blue!75!black,
  title={Agent (turns 2--3 -- redundant\_search, two hops on
  ``Jane Siberry birthplace'')}, boxrule=0.3mm, arc=2mm, breakable,
  fonttitle=\bfseries\small, fontupper=\small]
\emph{[Two consecutive turns querying ``Jane Siberry birthplace'' /
``Jane Siberry where born'' / ``Jane Siberry birthplace castle'';
both retrieve unrelated articles
(``Cornelia Jane Matthews Jordan'',
``John Roseberry Homestead'') and yield no relevant evidence.]}
\end{tcolorbox}

\begin{tcolorbox}[colback=blue!2!white, colframe=blue!75!black,
  title={Agent (turn 4 -- valid\_search, hit on hop $2$)}, boxrule=0.3mm,
  arc=2mm, breakable, fonttitle=\bfseries\small, fontupper=\small]
Based on the searches conducted, Jane Siberry, the performer of ``The
Speckless Sky'', hails from Canada. To find the castle located in her
birthplace, we should search for famous castles in Canada.\\[2pt]
\texttt{\textcolor{deepblue}{<tool\_call>}\,\{"query\_list":
["well-known castles in Canada"]\}\,\textcolor{deepblue}{</tool\_call>}}
\end{tcolorbox}

\begin{tcolorbox}[colback=blue!2!white, colframe=blue!75!black,
  title={Agent (turn 5 -- redundant\_search, on the right track but
  truncated)}, boxrule=0.3mm, arc=2mm, breakable, fonttitle=\bfseries\small,
  fontupper=\small]
Based on the information gathered, Jane Siberry, the performer of ``The
Speckless Sky'', is from Canada. Let's check if there are any famous
castles located in Canada:\\[2pt]
\texttt{\textcolor{deepblue}{<tool\_call>}\,\{"query\_list":
["famous castles in Canada", "Ch\^ateau Montebello characteristics"]\}\,\textcolor{deepblue}{</tool\_call>}}\\[3pt]
\emph{[Tool response Doc 1: ``Casa Loma (neighbourhood) is a neighbourhood
in the city of Toronto in Ontario, Canada, and is named after the
\textbf{famous castle}\ldots'' -- but the rollout terminates here at
the $5$-turn cap before the agent emits an
\texttt{<answer>\ldots</answer>} span.]}
\end{tcolorbox}

\subsection{Type $C$: miss-wrong ($H{=}0, C{=}0$)}
\label{app:example-C}

\begin{tcolorbox}[colback=gray!5!white, colframe=gray!60!black,
  title=Diagnosis, boxrule=0.3mm, arc=2mm, breakable,
  fonttitle=\bfseries\small, fontupper=\small]
\textbf{Question} (popqa): \emph{Who was the screenwriter for Abandoned?}\quad
\textbf{Reference}: \texttt{Elizabeth Sarnoff} / \texttt{Liz Sarnoff}.\quad
\textbf{Predicted}: \texttt{Zlatko Top\v{c}i\'c} (EM\,$=\!0$).\quad
\textbf{Pattern}: a name-disambiguation failure. The retriever returns
articles on \emph{The Abandoned (2010 film)} and \emph{Abandoned (1949
film)}, neither of which is the reference work, and the agent
confidently extracts the 2010 film's screenwriter without recognising
that the reference target is absent from the response. $\toolhit$ is
therefore $0$ at the rollout level. Under the routing matrix Type $C$
receives \emph{no} residual: the trajectory is a clean GRPO-handled
negative example, and any token-level direction we choose for either
segment would be unfounded.
\end{tcolorbox}

\begin{tcolorbox}[colback=blue!2!white, colframe=blue!75!black,
  title={Agent (turn 0 -- redundant\_search, no hit on the reference work)},
  boxrule=0.3mm, arc=2mm, breakable, fonttitle=\bfseries\small,
  fontupper=\ttfamily\small]
\textcolor{deepblue}{<tool\_call>}\\
\{"name": "search", "arguments":\\
\quad \{"query\_list": ["Who was the screenwriter for Abandoned",\\
\qquad\qquad\qquad\quad\ "movie Abandoned screenwriter"]\}\}\\
\textcolor{deepblue}{</tool\_call>}
\end{tcolorbox}

\begin{tcolorbox}[colback=orange!2!white, colframe=orange!75!black,
  title=Tool response (turn 0 -- wrong work), boxrule=0.3mm, arc=2mm,
  breakable, fonttitle=\bfseries\small, fontupper=\small]
\textbf{Doc 1 (Title: ``The Abandoned (2010 film)'')}: a 2010 Bosnian
drama film directed by Adis Bakra\v{c}\ldots and was \textbf{written by
Zlatko Top\v{c}i\'c}\ldots\\
\textbf{Doc 2 (Title: ``Abandoned (1949 film)'')}: based on an original
story by Irwin Gielgud\ldots\\
\textit{[Doc 3 omitted: another unrelated work. The reference target
(2010 ABC series \emph{Abandoned}, screenwriter Elizabeth Sarnoff) is
absent from the response, hence $\toolhit\!=\!0$.]}
\end{tcolorbox}

\begin{tcolorbox}[colback=blue!2!white, colframe=blue!75!black,
  title={Agent (turn 1 -- answer, \emph{wrong work})}, boxrule=0.3mm,
  arc=2mm, breakable, fonttitle=\bfseries\small, fontupper=\small]
Based on the search result, the screenwriter for the 2010 Bosnian drama
film ``The Abandoned'' (which seems to be the most relevant film with the
title ``Abandoned'') is Zlatko Top\v{c}i\'c.\\[2pt]
\textcolor{deeppurple}{<answer>}\,Zlatko Top\v{c}i\'c\,\textcolor{deeppurple}{</answer>}
\end{tcolorbox}

\subsection{Type $D$: miss-correct ($H{=}0, C{=}1$, unsupported correctness)}
\label{app:example-D}

\begin{tcolorbox}[colback=gray!5!white, colframe=gray!60!black,
  title=Diagnosis, boxrule=0.3mm, arc=2mm, breakable,
  fonttitle=\bfseries\small, fontupper=\small]
\textbf{Question} (nq): \emph{which way does the earth orbit the sun?}\quad
\textbf{Reference}: \texttt{counterclockwise}.\quad
\textbf{Predicted}: \texttt{counterclockwise} (EM\,$=\!1$, $\toolhit\!=\!0$).\quad
\textbf{Pattern}: the retriever returns descriptive passages on Earth's
orbit (eccentricity, perihelion, equinoxes) but \emph{none} of the
returned documents states the orbital direction explicitly, hence
$\toolhit\!=\!0$. The agent nevertheless commits to ``counterclockwise''
on parametric memory and happens to be correct.
This is the canonical Type-$D$ unsupported-correctness mode -- and the
ground for the methodology's split on retrievability: when the corpus \emph{does} contain supporting
evidence (Retrievable$\!=\!1$) the rollout receives a soft negative
residual to discourage parametric reliance; when the corpus does not, the
rollout is masked entirely.
\end{tcolorbox}

\begin{tcolorbox}[colback=blue!2!white, colframe=blue!75!black,
  title={Agent (turn 0 -- redundant\_search, no hit on direction)},
  boxrule=0.3mm, arc=2mm, breakable, fonttitle=\bfseries\small,
  fontupper=\ttfamily\small]
\textcolor{deepblue}{<tool\_call>}\\
\{"name": "search", "arguments":\\
\quad \{"query\_list":\\
\qquad ["which way does the earth orbit the sun",\\
\qquad\ "direction of earth's orbit around sun"]\}\}\\
\textcolor{deepblue}{</tool\_call>}
\end{tcolorbox}

\begin{tcolorbox}[colback=orange!2!white, colframe=orange!75!black,
  title=Tool response (turn 0 -- descriptive but no direction
  statement), boxrule=0.3mm, arc=2mm, breakable, fonttitle=\bfseries\small,
  fontupper=\small]
\textbf{Doc 1 (Title: ``Earth's orbit'')}: All celestial bodies in the
Solar System\ldots orbit around the Solar System's centre of mass\ldots
The average distance between the Earth and the Sun is 149.60 million km
(92.96 million mi), and one complete orbit takes 365.256 days
(1 sidereal year)\ldots eccentricity of 0.0167.\\
\textbf{Doc 2 (Title: ``Earth's orbit'')}: \ldots perihelion (periapsis --
nearest point to the Sun) on anywhere from January 2 to January 5,
the point of March equinox on March 19, 20, or 21\ldots\\
\textit{[None of the returned documents states that the orbit is
counterclockwise (or anti-clockwise / prograde) when viewed from the
ecliptic north; $\toolhit\!=\!0$.]}
\end{tcolorbox}

\begin{tcolorbox}[colback=blue!2!white, colframe=blue!75!black,
  title=Agent (turn 1 -- answer from parametric memory),
  boxrule=0.3mm, arc=2mm, breakable, fonttitle=\bfseries\small,
  fontupper=\small]
Based on the search results, Earth orbits the Sun in a counterclockwise
direction when viewed from above the Sun's north pole. This describes the
standard model of Earth's heliocentric orbit.\\[2pt]
\textcolor{deeppurple}{<answer>}\,counterclockwise\,\textcolor{deeppurple}{</answer>}
\end{tcolorbox}




\end{document}